\documentclass[lettersize,journal]{IEEEtran}
\usepackage{amsmath,amsfonts}
\usepackage{algorithmic}
\usepackage{algorithm}
\usepackage{array}
\usepackage[caption=false,font=normalsize,labelfont=sf,textfont=sf]{subfig}
\usepackage{textcomp}
\usepackage{stfloats}
\usepackage{url}
\usepackage{verbatim}
\usepackage{graphicx}
\usepackage{cite}
\usepackage[final,commandnameprefix=ifneeded]{changes}
\usepackage{cite}
\usepackage{amsmath,amssymb,amsfonts}
\usepackage{graphicx}
\usepackage{textcomp}
\usepackage{multirow}
\usepackage{ctable}
\usepackage{cleveref}
\usepackage{amsmath,amssymb}
\usepackage{xcolor}
\usepackage{url} 
\usepackage{caption}
\usepackage{subcaption}

\definechangesauthor[name={TIP2}, color=red]{2}

\newcommand{\blue}[1]{\textcolor{blue}{#1}}
\newcommand{\red}[1]{\textcolor{red}{#1}}

\newcommand{\mcc}[1]{\multicolumn{1}{c}{#1}} 

\newcommand*{\myrulefill}[3][]{%
  \makebox[#2]{#1%
    \leaders\hrule height \dimexpr.5ex+.2pt\relax depth \dimexpr -.5ex+.2pt\relax \hfill
    \enskip{#3}\enskip
    \leaders\hrule height \dimexpr.5ex+.2pt\relax depth \dimexpr -.5ex+.2pt\relax \hfill\kern0pt}
}

\begin{document}

\title{Uncertainty Propagation for Echocardiography Clinical Metric Estimation via Contour Sampling}

\author{Thierry~Judge, Olivier~Bernard, Nicolas~Duchateau, Woo-Jin~Cho~Kim, Alberto~Gomez, Arian~Beqiri, Agisilaos~Chartsias, and~Pierre-Marc~Jodoin
\thanks{T. Judge and P.-M. Jodoin are with the Department of Computer Science, University of Sherbrooke, Sherbrooke, QC, Canada (e-mail: thierry.judge@usherbrooke.ca).}
\thanks{T. Judge, N. Duchateau, and O. Bernard are with INSA, Université Lyon 1, CNRS UMR 5220, Inserm U1206, CREATIS, Villeurbanne, France.}
\thanks{N. Duchateau and O. Bernard are with Institut Universitaire de France (IUF).}
\thanks{ W.-J. Cho Kim, A. Gomez, A. Beqiri and  A. Chartsias are with Ultromics Ltd., Oxford, OX4 2SU, UK (A. Gomez, A. Beqiri and  A. Chartsias are employees and have shares in Ultromics Ltd.)}
\thanks{ A. Gomez is with King's College London, London, WC2R 2LS, UK}
}



\maketitle

\begin{abstract}
Echocardiography plays a fundamental role in the extraction of clinical parameters (e.g. left ventricular volume and ejection fraction) required to determine the presence and severity of heart-related conditions. When deploying automated techniques for computing these parameters, uncertainty estimation is \replaced{highly valuable}{crucial} for assessing their utility. Since clinical parameters are usually derived from segmentation maps, there is no clear path for converting pixel-wise uncertainty values into uncertainty estimates in the downstream clinical metric calculation. In this work, we propose a novel uncertainty estimation method based on contouring rather than segmentation. Our method explicitly predicts contour location uncertainty from which contour samples can be drawn. Finally, the sampled contours can be used to propagate uncertainty to clinical metrics. Our proposed method not only provides accurate uncertainty estimations for the task of contouring but also for the downstream clinical metrics on \replaced{one public and two private cardiac ultrasound datasets containing a total of 1264 patients}{two cardiac ultrasound datasets}.\footnote{Code will be rendered public upon acceptance of this paper.}
\end{abstract}


\section{Introduction}
\label{sec:introduction}

Ultrasound imaging is the predominant modality for the assessment of cardiac function\cite{Lancellotti2025}. One of the primary uses of echocardiography is the extraction of clinical metrics. For example, metrics such as left-ventricular volume and ejection fraction are widely used to diagnose heart disease\cite{Lang2015}. These metrics are measured by delineating the endocardial contours of the left ventricle at end-systole and end-diastole. The time-consuming nature of this task has stimulated interest in automatic methods. 

Deep learning methods have proved excellent at automating the extraction of clinical metrics from echocardiographic images\cite{Olaisen2023}. Such methods can be categorized in two classes: those where the metric values are regressed directly from the input image\cite{EchoGNN,EchoCoTr}; and those that first delineate the left ventricle and myocardium, either with a segmentation map or by contouring the endo- and epi-cardium, before computing the clinical metrics using these segmentations\cite{seg_metrics,seg_metrics2,Leclerc19}. In both cases, uncertainty estimation is of great importance in clinical practice. However, both approaches have respective shortcomings. 

On the one hand, uncertainty estimation of clinical metrics using regression models has been explored in~\cite{fimh_ef_unc,DEUE,dual_view_unc}. However, clinicians are reluctant to use non-intelligible black-box models, hindering their adoption in practice. Moreover, these models can only estimate metrics they were explicitly trained to predict. On the other hand, many works have addressed uncertainty estimation of cardiac segmentation\cite{card_mri_unc,judge_crisp,unc_2d_echo}. However, to our knowledge, no work has investigated the influence of this uncertainty on the derived clinical metrics. 

The main contribution of this paper is a generalized framework to propagate uncertainty from input images to segmentation and to the derived clinical metrics while, at the same time, exhibiting segmentation accuracies on par with SOTA methods. Unlike conventional methods yielding pixel-wise segmentation and uncertainty maps, our contour-based approach, particularly suited to clinical settings where user interactions remain critical,  predicts aleatoric uncertainty directly for points defining the left ventricle. This method, combined with shape uncertainty estimates, enables sampling plausible and temporally consistent contours and gives rise to a robust end-to-end pipeline for automatic clinical metric estimation with intelligible uncertainty.

\begin{figure*}[tp]
\centering
\includegraphics[width= \textwidth]{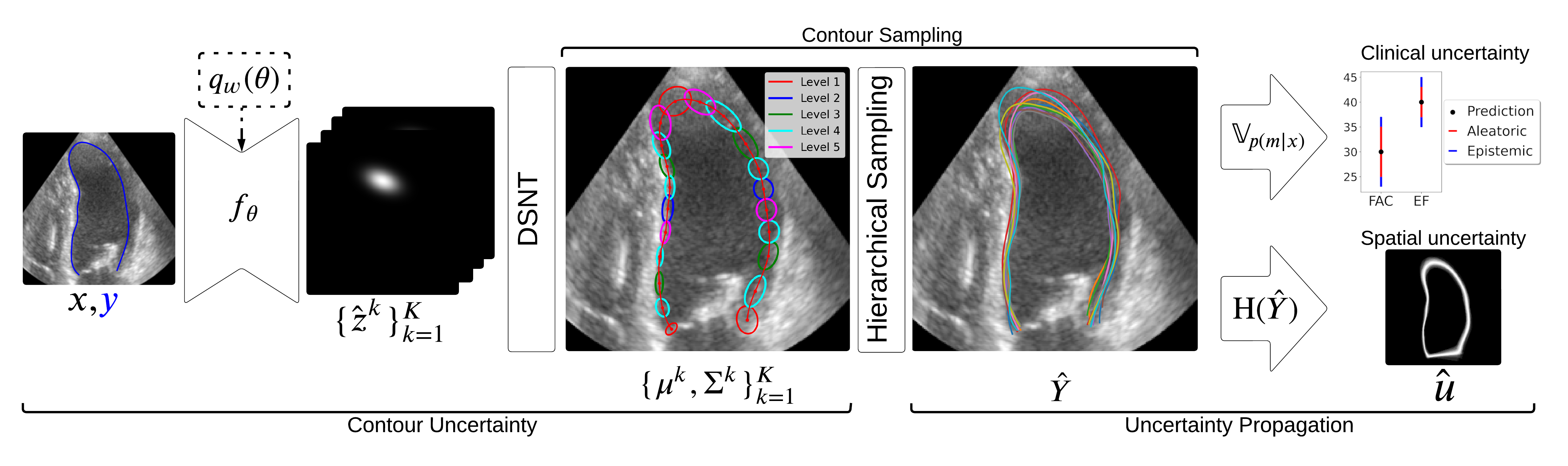}
\caption{Schematic of the CASUS framework illustrating the three core components: (1) prediction of contour point uncertainty, (2) a contour sampling mechanism, and (3) propagation of sampled contour uncertainties to compute clinical metrics.}
\label{fig:pipeline}
\end{figure*}


\section{Related Work}
\label{sec:related_work}

\subsection{Echocardiographic Segmentation}

Deep neural networks trained on publicly available datasets such as CAMUS~\cite{Leclerc19} and EchoNet-Dynamic~\cite{Ouyang2020_echonet} have reached a performance within inter-user variability, typically using architectures based on U-Net \cite{unet}. Since then, many different strategies such as echocardiography-specific data augmentation \cite{GUDU} and pyramid attention networks \cite{pyramid_att_seg} have improved U-Net results. 
The recent publication of nnU-Net has significantly enhanced segmentation outcomes across various tasks \cite{isensee2021nnu} and, when applied to the CAMUS dataset, outperformed all prior methods~\cite{camus_nnUnet}.

Other work has addressed key-point localization~\cite{thaler_heatmaps,Schobs_heatmaps} with dedicated approaches for cardiac~\cite{lv3ch-dsnt} and lung\cite{Gaggion_2022} segmentation being proposed. This formulation reduces anatomical implausibilities and is more similar to clinical tools where keypoints are used to define contours in commercial software such as EchoPac by GE Healthcare\footnote{\url{https://www.gehealthcare.com/products/ultrasound/vivid/echopac}}, which includes semi-automatic and manual delineation tools.

Regardless of the segmentation technique, a key objective following segmentation is the derivation of clinical metrics. A standard metric in echocardiography is the left ventricular area. A more accurate evaluation of the ventricular cavity size involves determining the ventricular volume using Simpson's biplane method~\cite{2020LeftVE}, utilizing the segmentation obtained from the apical 4-chamber (A4C) and apical 2-chamber (A2C) views. Additionally, temporal metrics are of significant interest. Examining the ratio of heart size between end-systole (ES) and end-diastole (ED) can provide indications of certain cardiac pathologies. The fractional area change (FAC) is defined as the difference between the end-diastolic area and the end-systolic area, divided by the end-diastolic area. Ejection fraction (EF) is calculated similarly, albeit with volume measurements.

\subsection{Uncertainty Estimation}

Although numerous studies have proposed new models to enhance the segmentation accuracy, relatively few have addressed the predictive uncertainty of these models. Accounting for uncertainty is crucial, as although neural networks can achieve human-level performance, they can also make occasional critical errors.

Uncertainty is usually classified into two categories: {\em aleatoric} and {\em epistemic}~\cite{bayesianCV}. Aleatoric uncertainty is the uncertainty present within the data itself. In medical imaging, aleatoric uncertainty is caused by the inherent ambiguity of images, which for instance, can occur due to noisy images or labels.
Epistemic uncertainty is the uncertainty associated with the model and is caused by the limited size of the training dataset. While epistemic uncertainty can be reduced by acquiring more data, aleatoric uncertainty is irreducible.

One prominent approach for modeling epistemic uncertainty involves characterizing the distribution of model weights, $\theta$, conditioned on the data $\mathcal{D}$. This posterior distribution over weights, $p(\theta|\mathcal{D})$, is generally intractable to compute analytically for neural networks. Variational inference offers a method to approximate the true posterior using a variational distribution, $q_w(\theta)$, in Bayesian neural networks~\cite{pmlr-v37-blundell15}. However, this approach necessitates significant modifications to standard training procedures. Alternative methods, such as Monte-Carlo (MC) dropout, have demonstrated effectiveness as a practical approximation for Bayesian neural networks, allowing for uncertainty estimation without altering the training pipeline, and have achieved reliable results~\cite{Gal_2016_bayesNN, Gal_2015_BayesianCNN}. On the other hand, a frequentist approach with ensemble methods leverages the variability of independently trained networks—each trained on different data subsets or with distinct hyperparameters—to capture uncertainty~\cite{ensembles}.

While epistemic uncertainty methods usually do not require any data assumptions, this is not the case for aleatoric uncertainty methods. 
Most commonly, probabilistic approaches that model a parameterized distribution $p_\theta(y|x)$ over the output $y$ given the input $x$ are employed in aleatoric uncertainty methods. For instance, in the case of univariate regression, the mean and variance of a Gaussian distribution are learned by maximizing the negative log-likelihood with respect to both parameters. This can be extended to the multivariate normal~\cite{multivariate_uncertainty} or with different distributions such as the skew-normal distribution~\cite{miccai2023_contouring}. In the case of classification, uncertainty is approximated by modeling a Gaussian distribution across every output logit~\cite{bayesianCV}.

Aleatoric uncertainty can also be predicted with non-probabilistic methods. Methods that use test-time augmentation apply random data transformations, perform inference and aggregate the outputs to model uncertainty~\cite{ayhan2018_tta}.

While modeling both aleatoric and epistemic uncertainty is relevant to understanding the origin of the prediction ambiguity, it is also important to model the full predictive uncertainty. If the aleatoric uncertainty is captured by a model $p_\theta(y|x)$ and the epistemic uncertainty is modeled by $q_w(\theta)$, the predictive uncertainty is given by the law of total variance~\cite{bayesianCV}
\begin{equation}
    \mathbb{V}_{p(y|x)} = \mathbb{E}_{q_w(\theta)}\Big[\mathbb{V}_{p_\theta(y|x)}[y]\Big] + \mathbb{V}_{q_w(\theta)}\Big[\mathbb{E}_{p_\theta(y|x)}[y]\Big].
\end{equation}
\subsubsection{Segmentation uncertainty}

In many cases, segmentation uncertainty is tackled similarly to classification. Epistemic uncertainty methods, such as MC dropout and ensembles, do not require any changes as multiple forward passes generate varying segmentation maps. In the case of aleatoric uncertainty, a simple way is to consider the prediction for each pixel as an independent classifier and estimate aleatoric uncertainty by modeling the logits of each pixel with a Gaussian~\cite{bayesianCV}.

Pixel-wise aleatoric uncertainty often performs poorly since independent pixel distributions make plausible segmentation samples unlikely. For this reason, methods such as Probabilistic U-Net \cite{prob-unet} and Phi-Seg \cite{phiseg} use a variational autoencoder coupled with a segmentation network to learn a distribution of shapes in the latent space from which plausible samples can be drawn. Similarly, stochastic segmentation networks learn a low-rank multivariate Gaussian representation of pixel distributions, enabling efficient training despite the large covariance matrix, and outperform previous methods~\cite{stochastic_seg_net}.

Finally, test-time augmentation can be extended to segmentation by applying transforms to the input image and applying the inverse transform on the output segmentation~\cite{WANG2019_tta_seg}.

\subsubsection{Uncertainty in echocardiographic image analysis}

Numerous approaches have been developed to estimate uncertainty in echocardiography, usually focusing on segmentation or clinical metric prediction. For the former task, techniques like ensembles and test-time augmentation have been shown to improve left-ventricle segmentation performance (as measured by Dice scores) by excluding the most uncertain samples~\cite{unc_2d_echo}. Zang \textit{et al} combined metrics derived from ensembles with convexity metrics, showing that this improved detection of samples with large errors~\cite{Zhang2022}. Judge \textit{et al.} took a post-hoc approach, predicting segmentation uncertainty by encoding prior information in a shared latent space~\cite{judge_crisp}. For the latter task, Behnami \textit{et al.}~\cite{dual_view_unc} proposed a multitask learning approach to estimate EF alongside its uncertainty for both the A2C and A4C views, using Gaussian modeling. Their work also included a classification-based prediction of EF, categorizing cardiac function into four classes from normal to severe dysfunction. In a similar vein, Kazemi Esfeh \textit{et al.} introduced a sampling-free ensemble method to estimate epistemic uncertainty, specifically targeting EF regression with a delta ensemble approach~\cite{DEUE}. 

Although these methods may propose robust prediction and uncertainty measures, they lack the intelligibility of clinical metric prediction extracted from segmentation maps. Jafari \textit{et al.}~\cite{Jafari_bayesian_ef_uncertainty} proposed a method to estimate segmentation uncertainty using MC dropout. They then define an uncertainty measure based on this uncertainty map and use it to filter samples showing an increased correlation between predicted ejection fraction values and ground truth values. While this is similar to our work, there is no explicit uncertainty estimation for the ejection fraction, as their method only filters out samples with uncertainty above a certain threshold. \added{A recent work by Zhao et al.~\cite{Zhao_aleatoric} addresses uncertainty propagation with a similar contour-based approach, applied to cardiac MRI segmentation. Our preprint, publicly available since February 2025~\cite{judgeUncertaintyPropagationEchocardiography2025a}, was developed independently and applies this idea to the distinct context of ultrasound-based cardiac segmentation.}



In summary, existing methods have failed to connect uncertainty for segmentation and the resulting clinical metrics. This paper addresses this limitation by providing a method to quantify and propagate uncertainty from the predicted shape to the derived metrics in an interpretable way.

\section{Method}

A standard segmentation pipeline for estimating clinical metrics in echocardiographic images requires a dataset consisting of $N$ pairs $\{x_n, y_n\}_{n=1}^N$, where each pair is an image $x_n \in \mathbb{R}^{H \times W}$ of height $H$ and width $W$, and corresponding ground truth segmentation map of $C$ classes $y_n \in \{0,1\}^{C \times H \times W}$. This dataset is used to train a segmentation network $f^{seg}_\theta$. For a given input image, $x_n$, the segmentation network outputs a segmentation map $\hat{y}_n = f^{seg}_\theta(x_n)$. Given one (or many) segmentation maps $\hat{y}_n \in \{0,1\}^{C \times H \times W}$, a clinical metric $F(\cdot)$ is computed, e.g the area of a segmentation map or the left ventricle ejection fraction given the 2CH and 4CH view segmentation for the ES and ED frames. 

We aim to estimate the uncertainty of clinical metrics by propagating the uncertainty from segmentation maps or contours, the latter of which is the focus of this paper. 
Our method, Contour Aleatoric Shape Uncertainty Sampling (CASUS), is illustrated in \Cref{fig:pipeline} and contains three main components. The first is the contour uncertainty prediction, where a neural network is trained to predict points defining the contour of the left-ventricle and their uncertainty. The second component is a hierarchical contour sampling algorithm that allows for the sampling of plausible contours from independent point predictions. This sampling process relies on a conditional shape model that allows the modeling of shape uncertainty. Finally, the third step propagates the uncertainty from the contour uncertainty prediction to the clinical metrics.

\subsection{Contour-based Aleatoric Uncertainty Modeling}

Contrary to most previous work in uncertainty estimation for segmentation, we model the uncertainty of contour points rather than the pixel-wise uncertainty. We advocate that this is more appropriate for left-ventricle segmentation as there is no uncertainty around the presence of the left-ventricle in the image, but rather uncertainty regarding its position. This aspect of our work focuses on estimating aleatoric uncertainty, as epistemic uncertainty methods are well-established and usually task-agnostic. Our method can be combined with epistemic uncertainty methods, as shown in \Cref{sec:uncertainty_prop}. 

The prediction of a contour and its associated uncertainty begins with a training dataset made of $\mathcal{D}=\{x_n, s_n\}_{n=1}^N$, where $s_n$ is a series of $K$ ordered points $s^k_n$ drawn by an expert or extracted from a segmentation maps.  With $\mathcal{D}$, a contouring neural network $f_\theta^c(\cdot)$ is trained to predict $K$ heatmaps $Z^k \in \mathbb{R}^{H \times W}$, each associated to a control point, as illustrated on the left of~\Cref{fig:pipeline}.  Each heatmap contains a blurry white spot whose center corresponds to the position of a control point (variable $\mu$) and whose halo size corresponds to the uncertainty of that point (variable $\Sigma$).  The heatmaps are normalized in 2D to obtain $\hat{Z^k}$, such that it represents a probability density function for the coordinates $c^k$ of a given point $k$ in the shape. 

We build upon the differentiable spatial to numerical transform (DSNT) \cite{nibali2018numerical} to extract the parameters of a 2D Gaussian from the heatmap. To extract the mean and variance of the heatmap for each point, two coordinate maps $\mathbf{I} \in \mathbb{R}^{H \times W}$ and $\mathbf{J} \in \mathbb{R}^{H \times W}$, where $\mathbf{I}_{i,j} = \frac{2j - (W+1)}{W}$ and $\mathbf{J}_{i,j} = \frac{2i - (H+1)}{H}$ are used. The mean and variance can be extracted as follows,
\begin{equation}
     \hat\mu^k = \mathbb{E}[c^k] = \Big[ \langle\hat{Z^k}, \textbf{I}\rangle_F, \langle\hat{Z^k}, \textbf{J}\rangle_F\Big] \in \mathbb{R}^2,
     \label{eq:mu}
\end{equation}
\begin{equation}
     \label{eq:var1}
\begin{split}
\mathbb{V}[c^{k_j}]
& = E[(c^{k_j} - E[c^{k_j}])^2) \\
& = \langle\hat{Z}^k, (\textbf{I} - \mu^{k_j}) \odot (\textbf{I} - \mu^{k_j})\rangle_F \in \mathbb{R},
\end{split}
\end{equation}
where $\langle\cdot, \cdot\rangle_F$ is the Frobenius inner product, $\odot$ corresponds to the Hadamard product, and $\mathbb{V}[c^{k_i}]$ is computed similarly. The covariance can also be extracted from the heatmap, 
\begin{equation}
\begin{split}
\mathbb{C}[c^k]
& = E[(c^{k_j} - E[c^{k_j}])(c^{k_i} - E[c^{k_i}]) \\
& = \langle\hat{Z}, (\textbf{I} - \mu^{k_j}) \odot (\textbf{J} - \mu^{k_i})\rangle_F \in \mathbb{R}. 
\end{split}
\end{equation}

By combining the computed variance and covariance values, the covariance matrix can be constructed and a 2-dimensional Gaussian distribution can be modeled.

\begin{equation}
\hat\Sigma^k =    \begin{bmatrix}
            \mathbb{V}[c^{k_j}] & \mathbb{C}[c^k]\\
            \mathbb{C}[c^k] & \mathbb{V}[c^{k_i}]
            \end{bmatrix} 
            \in \mathbb{R}^{2 \times 2}
\end{equation}

The model thus predicts a tuple $(\hat\mu, \hat\Sigma)$ for each image $x$, with $\hat\mu \in \mathbb{R}^{K \times 2}$ and $\hat\Sigma \in \mathbb{R}^{K \times 2 \times 2}$ (c.f. Fig.~\ref{fig:pipeline} middle). To train the network, the negative log-likelihood of the Gaussian is used for each point, 

\begin{equation}
    \mathcal{L} = \frac{1}{NK} \sum_{n=1}^N \sum_{k=1}^{K} \frac{1}{2}\log{|\hat\Sigma_{n}^{k}|} + \frac{1}{2}(\hat\mu_{n}^{k} - s_{n}^{k})^T (\hat\Sigma_{n}^{k})^{-1} (\hat\mu_{n}^{k} - s_{n}^{k}) \nonumber
\end{equation}

\subsection{Shape Uncertainty Modeling}

\begin{figure}[tp]
\centering
\includegraphics[width=0.99\linewidth]{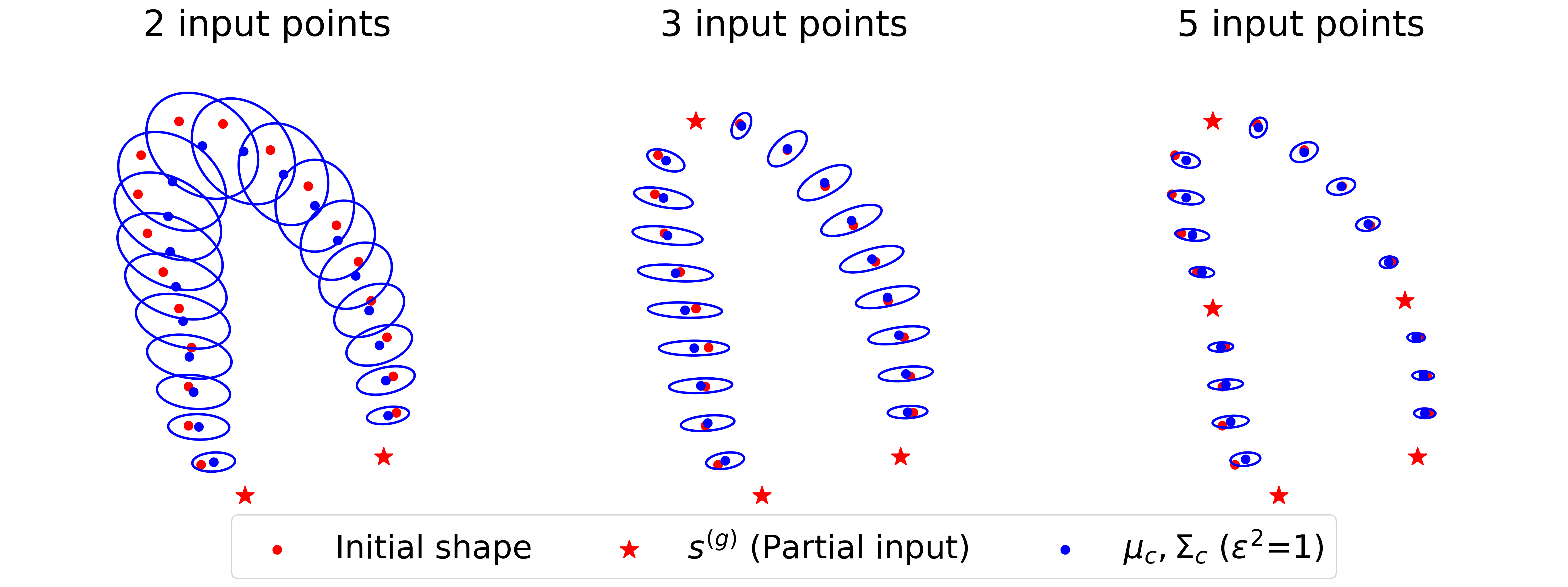}
\caption{Example of the posterior shape model output for various partial inputs of a given left ventricle shape. }
\label{fig:psm_example}
\end{figure}

\begin{figure*}[tp]
  \centering
  \subfloat[Level 2 sampling\label{fig:sampling_level}]{
    \includegraphics[width=0.24\textwidth]{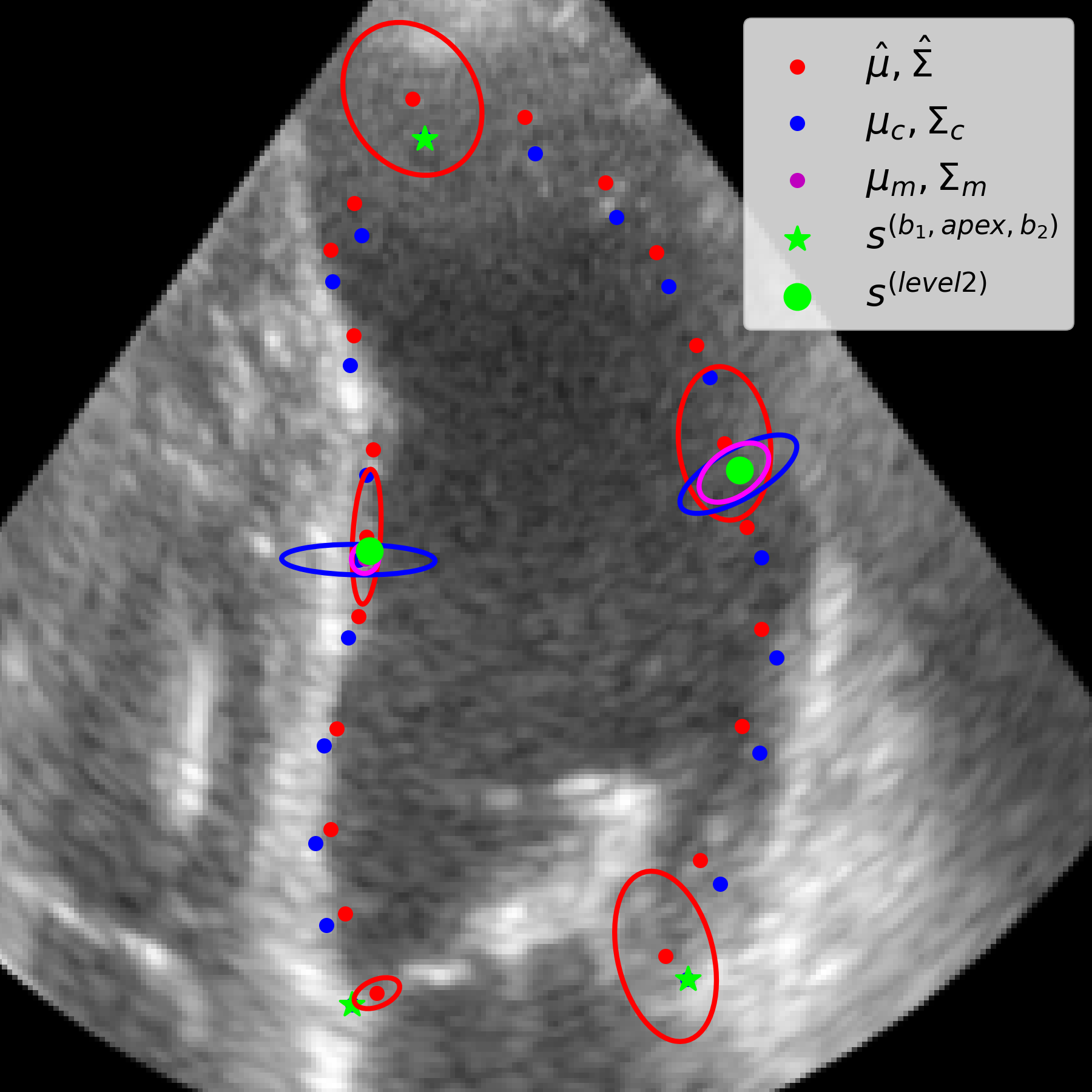}
  }\hfill
  \subfloat[Post sampling distribution\label{fig:post_dist}]{
    \includegraphics[width=0.24\textwidth]{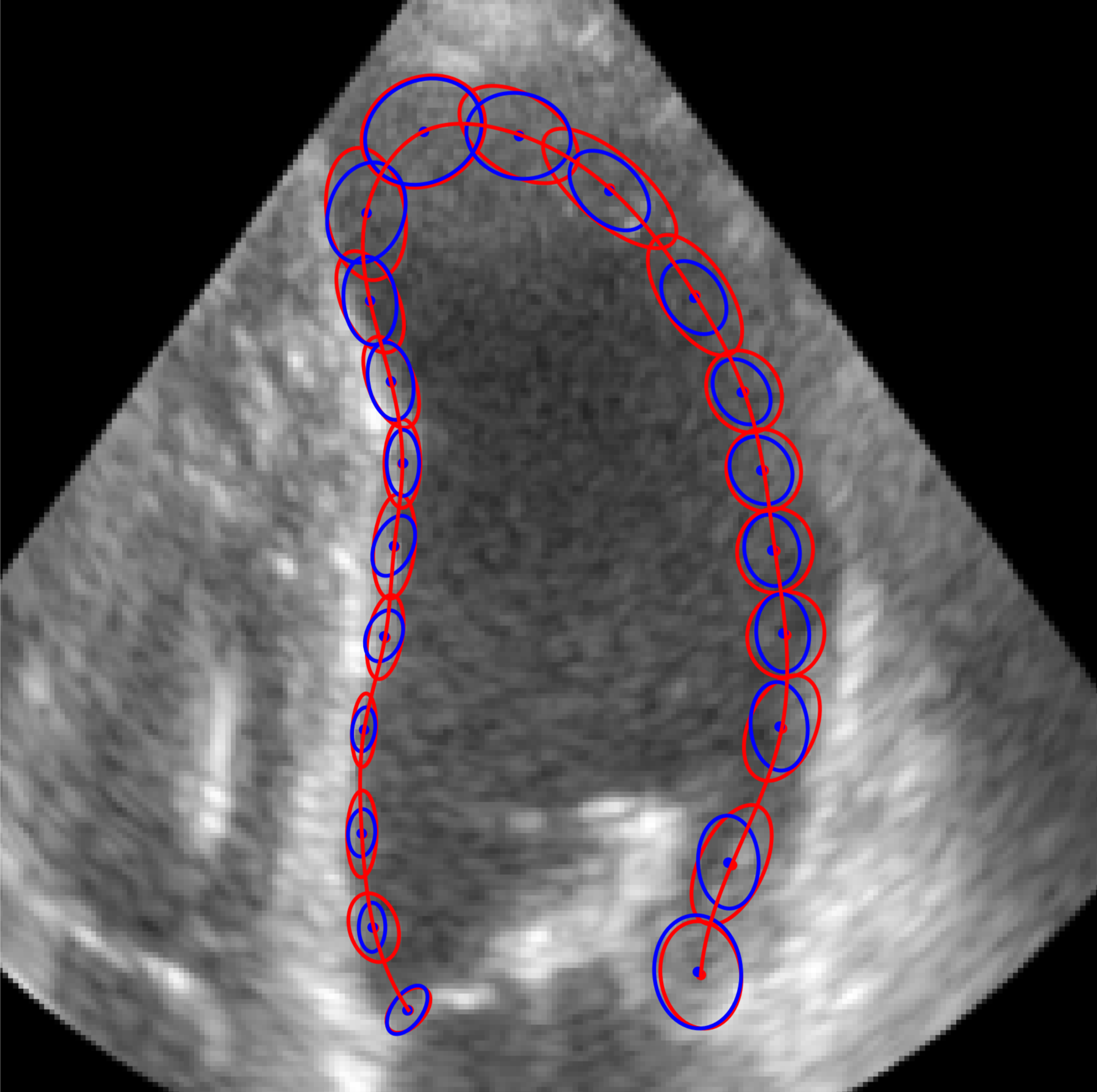}
  }\hfill
  \subfloat[Temporal sampling\label{fig:seq_sampling}]{
    \includegraphics[width=0.47\textwidth]{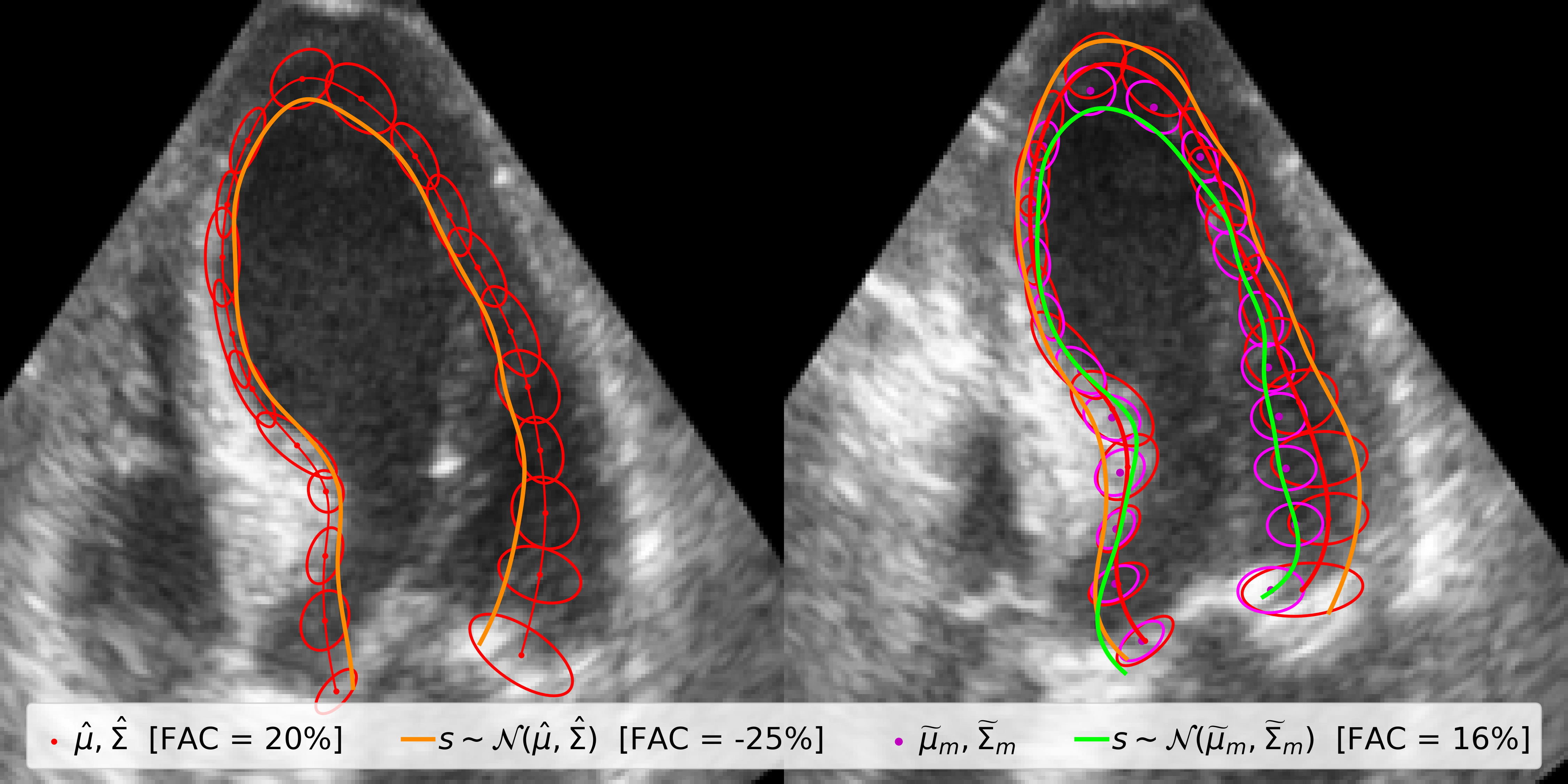}
  }
  \caption{(a) Example of the sampling process. Given the initial points (green stars), the posterior shape model output (blue) is merged with the predicted distribution (red) to obtain a distribution (magenta) from which points are sampled (green). (b) The predicted contour and uncertainty distributions in red, and the points sampled with the PSM sampler 
  in blue. (c) Temporal consistent sampling. The orange curve shows an example of the failure case of independent sampling where the ED area (left) is smaller than the ES area (right), resulting in a negative FAC greatly differing from the mean prediction. The time-consistent sampling produces distributions (magenta) that are consistent with the ED contour, resulting in a good contour for ES (green).}
\label{fig:sampling_method}
\end{figure*}

While the previous formulation allows predicting one contour, the uncertainty propagation to clinical metrics requires multiple contours, which can be retrieved with stochastic sampling of $\{\hat{\mu^k},\hat{\Sigma^k}\}_{k=1}^K$. However, sampling each individual distribution $\{\hat{\mu^k},\hat{\Sigma^k}\}$ does not provide valid contours as the inter-point covariance is not modeled. The entire shape distribution requires predicting a $2K \times 2K$ covariance, which is numerically unstable. We thus propose using a shape model to model the shape uncertainty between points.

We do this with a statistical shape model that models the uncertainty of points given the position of other points. To define the conditional distribution we use a posterior shape model (PSM) based on principal component analysis (PCA), which predicts conditional distributions given partial data~\cite{PSM_ALBRECHT2013}.  We start by introducing the PCA notation used in the PSM. 

\noindent\textbf{PCA:} Given a set of $N$ shapes $\textbf{S}=\{s_1, s_2, ..., s_N\}$, each containing $K$ 2D points, $s_n \in \mathbb{R}^{2K}$, i.e. the training set used for contouring, the mean, $\bar\mu \in \mathbb{R}^{2K}$ and covariance matrix, $\bar\Sigma \in \mathbb{R}^{2K \times 2K}$ of these $N$ shapes can be computed as follows, 
\begin{equation}
    \bar{\mu} = \frac{1}{N}\sum_{i=n}^N s_n, \quad \bar{\Sigma} = \frac{1}{N}\sum_{n=1}^N (s_n - \bar\mu)(s_n - \bar{\mu})^T.
\end{equation}

By performing the eigenvalue decomposition of the covariance matrix, $\bar\Sigma = U \Lambda^2 U^T$, all shapes can be represented by 
\begin{equation}
\label{eq:pca}
    s = s(\alpha) = \bar\mu + U\Lambda \alpha = \bar{\mu} + Q \alpha. 
\end{equation}

If $\alpha \sim \mathcal{N}(0, I)$, the shapes $s(\alpha) =  \bar{\mu} + Q \alpha$ are distributed according to $\mathcal{N}(\bar{\mu}, \bar{\Sigma})$. 

\noindent\textbf{PSM:} The goal of the posterior shape model is to produce a conditional distribution given partial data.  More formally, given partial information, i.e. a selected subset of points $s^{(g)} \in \mathbb{R}^{2q}$, where $q < K$, the goal is to estimate $p(s|s^{(g)})$. Per \cite{PSM_ALBRECHT2013}, the conditional distribution is a multivariate Gaussian,  
\begin{equation}
    p(s|s^{(g)}) = \mathcal{N}(\mu_c, \Sigma_c).
\end{equation}
The mean and covariance matrix are given by  
\begin{equation}
    \mu_c = \bar{\mu} + Q(Q_g^TQ_g + \epsilon^2I_{2K})^{-1}Q_g^T(s^{g} - \mu_g) \in \mathbb{R}^{2K},
\label{eq:psm_mu}
\end{equation}
\begin{equation}
    \Sigma_c = \epsilon^2Q(Q_g^TQ_g + \epsilon^2I_{2K})^{-1}Q_g^T  \in \mathbb{R}^{2K \times 2K},
\label{eq:psm_simga}
\end{equation}
where $\mu_g \in \mathbb{R}^{2q}$ is the sub-vector of $\bar{\mu}$ for corresponding entries of $s_g$, $Q_g \in \mathbb{R}^{2q\times 2K}$ is the sub-matrix of rows of $Q$ for corresponding entries of $s^{(g)}$ and $\epsilon^2$ is a slack term to account for variations in data from the original training set. 

While the PSM is defined to use the mean $\bar\mu$ of the shape set to compute $Q$ and in \Cref{eq:psm_mu} and \Cref{eq:psm_simga}, we instead use $\hat\mu$ as this allows the PSM to model the possible variations in the dataset, and therefore the shape uncertainty, with respect to the predicted shape. We will refer to the matrix $Q$ computed with respect to $\hat\mu$ as $\hat{Q}$. We will denote the operation of computing the conditional distribution given a partial input as follows, 
\begin{equation}
        \mu_c, \Sigma_c = \text{PSM}(s^{(g)}, \hat{\mu}, \hat{Q}).
\end{equation}

An example of the PSM on the shape of the left ventricle is illustrated in ~\Cref{fig:psm_example} showing that points close to the partial input have low variance. Also, as the size of the partial input increases, the uncertainty over the remaining points' positions is lower therefore restricting the possible positions.

\subsection{Aleatoric and Shape Uncertainty Fusion}

Given appropriate methods for modeling both the aleatoric contour uncertainty and the shape uncertainty, we can propose an algorithm to sample multiple contours. We do this by defining a hierarchical sampling algorithm that fuses both the contour uncertainty and shape uncertainty.

By the very nature of the data and labeling process of the left-ventricle shape, it can be assumed that only the apex and the two basal points are independent. These points are tied to well-defined anatomical landmarks: the basal points are located at the base of the mitral valve, and the apex is the farthest point from these points, located at the tip of the left ventricle. The other points in the shape are present to define the contour along the endocardial border, but do not have a true anatomical landmark in the image. 

Our hierarchical sampling process, therefore, begins by sampling both basal points and apex points independently. Their position is then used to define a conditional distribution that is used to update the neural network's prediction. 

To sample a shape $s$ given the outputs of the uncertainty contouring model, $\{ \hat{\mu}^k, \hat{\Sigma}^k \}_{k=1}^K$, the apex is initially sampled,
\begin{equation}
    s^{(\text{apex})} \sim  \mathcal{N}(\hat\mu^{(\text{apex})}, \hat\Sigma^{(\text{apex})}).
\end{equation}
The same is done for both basal points ($\text{b}_1$ and $\text{b}_2$). Given this partial input, the parameters of the posterior shape model distribution can be predicted 
\begin{equation}
        \mu_c, \Sigma_c = \text{PSM}(s^{(\text{b}_1, \text{apex}, \text{b}_2)}, \hat{\mu}, \hat{Q}).
\end{equation}
To obtain a distribution conditioned on both the original model prediction and the sampled points, we fuse the contour uncertainty distributions and the shape uncertainty given by the posterior shape model distribution. We merge both distributions by multiplying them, as is done when combining a likelihood and prior term. The multiplication of the two multivariate Gaussians is defined as follows
\begin{multline}
    \mu_m^{k} = \hat\Sigma^{k} (\hat\Sigma^{k} + \Sigma_c^{k})^{-1} \mu_c^{k} + \Sigma_c^{k} (\hat\Sigma^{k} + \Sigma_c^{k})^{-1} \hat\mu^{k}, 
    \label{eq:merge_mu}
\end{multline}
\begin{equation}
    \Sigma_m^{k} = \hat\Sigma^{k} (\hat\Sigma^{k} + \Sigma_c^{k})^{-1} \Sigma_c^{k},
        \label{eq:merge_sigma}
\end{equation}
where $\Sigma_c^{k}$ is $2 \times 2$ sub-matrix extracted from $\Sigma_c$ along the diagonal at the indices corresponding to point $k$. With a new conditioned distribution for each point, we sample the midpoints between the previously sampled points and repeat the process until all points are sampled at each hierarchical level (shown in \Cref{fig:pipeline}).

The second sampling level of the hierarchical sampling process is illustrated in \Cref{fig:sampling_level}. In \Cref{fig:post_dist}, the final point distribution is compared with the original predicted distribution, showing that the PSM sampler does not alter the point distribution except for the points sampled last.

\subsubsection{Temporal consistent sampling}
\label{sec:temporal_sampling}

One critical consideration in cardiac ultrasound is the temporal aspect, as many clinical parameters are derived from segmented data at distinct time points. As our contouring method makes independent predictions for the ES and ED frames, as is the case for most segmentation methods, there is a risk of obtaining sets of contours that lack physiological plausibility. For instance, sampling a contour inside the mean contour during end-diastole (when the left ventricle is at its largest) and another contour outside the mean contour during end-systole (when the left ventricle is at its smallest) could result in a negative FAC, which contradicts physiological norms.

To reduce the occurrence of implausible samples while sampling, we use the posterior shape model to restrict the distribution of the contour sampled at the second time instant. To do so, we consider the set of shapes for ES and ED instants $\widetilde{\mathbf{S}} \in \mathbb{R}^{N \times 4K}$. We compute the mean $\widetilde{\mu} \in \mathbb{R}^{4K}$ and covariance matrix $\widetilde{\Sigma} \in \mathbb{R}^{4K \times 4K}$ of this set, from which we can extract the PCA parameter $\widetilde{Q}$. We then sample the first contour from the ED predictions with the hierarchical sampling process. We compute the posterior shape model parameters using, \Cref{eq:psm_mu} and \Cref{eq:psm_simga}, to obtain $\widetilde{\mu}_c$ and $\widetilde{\Sigma}_c$. We merge the predicted distribution for the ES with the PSM output with \Cref{eq:merge_mu} and \Cref{eq:merge_sigma} to obtain $\widetilde{\mu}_m$ and $\widetilde{\Sigma}_m$ from which a contour is derived with hierarchical sampling. In practice, the first sampled contour is randomly chosen between the ES and ED frames. An example of the sequential sampling process is shown in \Cref{fig:seq_sampling}.

\begin{figure*}[tp]
\centering
\includegraphics[width=\textwidth]{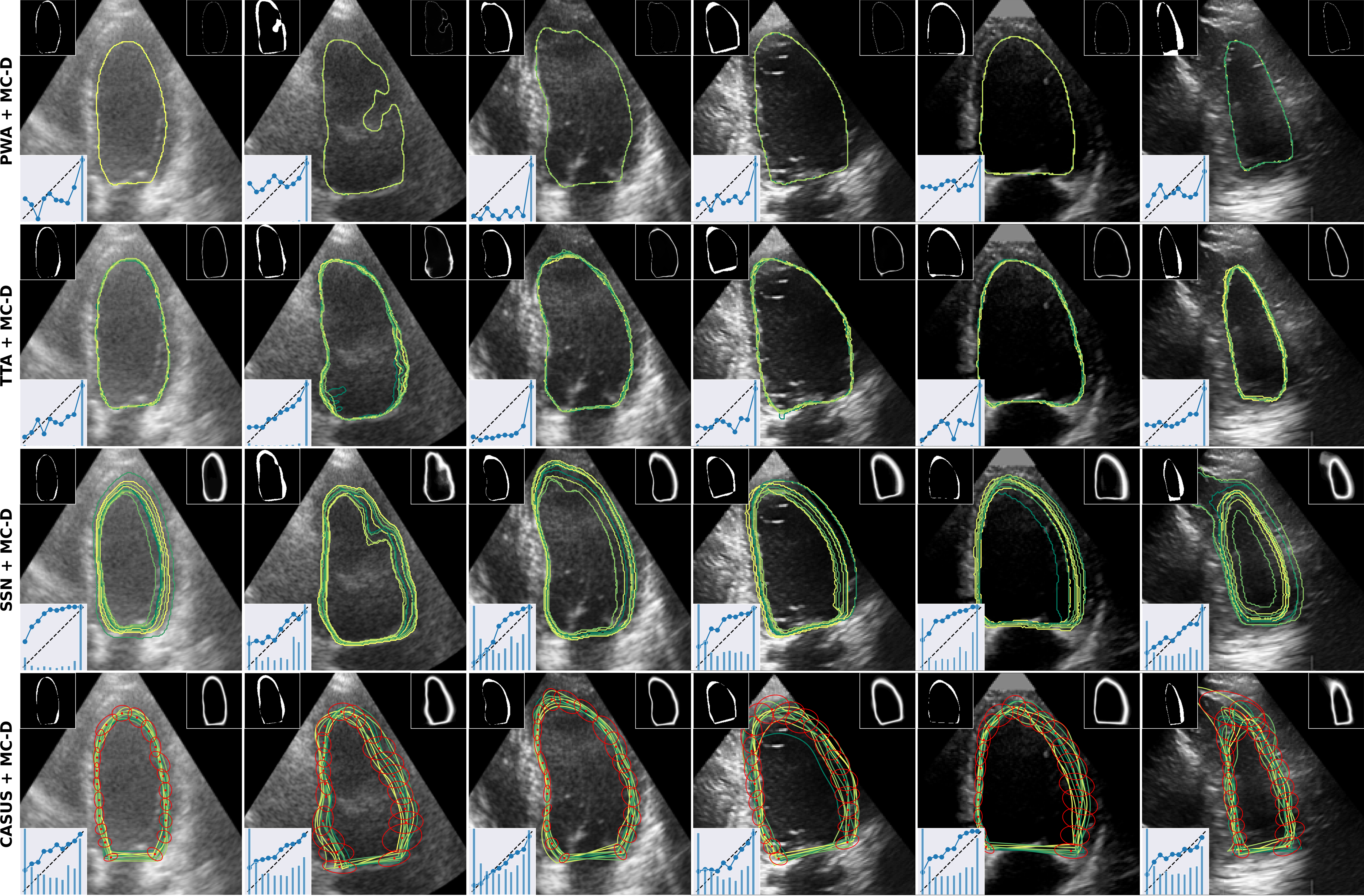}
\caption{Example predictions from different methods. Each row corresponds to a different method and each column a specific image. In each image, 10 samples are shown. The top left and right sub-images are the pixel-wise error and uncertainty maps, respectively. For each image, the pixel-wise calibration reliability diagram is plotted in the bottom left corner. The histogram indicates the number of samples in each bin. The last row corresponds to the CASUS method, where the red ellipses correspond to 95\% confidence intervals for the location of each point. }
\label{fig:samples}
\end{figure*}

\subsection{Uncertainty Propagation}
\label{sec:uncertainty_prop}

In the previous two subsections, we presented a contour aleatoric uncertainty estimation method.  This method allows to model the distribution of plausible samples for a given image which we denote as $p_\theta(s|x)$, where $s$ is a left-ventricle shape and $x$ is the input ultrasound image. 
To allow for this section of the paper to be compared with standard segmentation methods, we consider the segmentation map obtained by filling the inside of a contour $s$ and denote it $y$. Assuming we have an approximated posterior distribution for the neural network weights, $q_w(\theta)$, we can propagate both the aleatoric and epistemic segmentation uncertainty to the clinical metrics. 

If the value of a certain clinical metric $F$ is given by $m = F(y)$ where $y$ is a segmentation map, the predictive uncertainty, i.e. the sum of the aleatoric and epistemic uncertainty, of $m$ is given by

\begin{equation}
        \mathbb{V}_{p(m|x)} = \mathbb{E}_{q_w(\theta)}\Big[\mathbb{V}_{p_\theta(y|x)}[F(y)]\Big] + \mathbb{V}_{q_w(\theta)}\Big[\mathbb{E}_{p_\theta(y|x)}[F(y)]\Big]. \nonumber
\end{equation}

In the case where only epistemic uncertainty is computed (i.e. $p_\theta(y|x)$ is a point estimate), the first term will be zero. Conversely, if $q_w(\theta)$ is a point estimate, no epistemic uncertainty is estimated. 

To compute this uncertainty, we use a Monte Carlo framework. We define $T_a$ to be the number of samples to estimate the aleatoric uncertainty and $T_e$ to be the number of epistemic samples. We compute a series of predictions
\begin{equation}
    \hat{Y} = \{(\hat{y}^1_1, \hat{y}^1_2,...,\hat{y}^1_{T_a}), ..., (\hat{y}^{T_e}_1, \hat{y}^{T_e}_2,...,\hat{y}^{T_e}_{T_a})\},
\end{equation}
where $\hat{y}^i_j \sim p_{\theta^t}(y|x)$ with $\theta^t \sim q_w(\theta)$. For each prediction in $\hat{Y}$, a given clinical metric $F$ can be computed to obtain 
\begin{equation}
    M = \{(m^1_1, m^1_2,...,m^1_{T_a}), ..., (m^{T_e}_1, m^{T_e}_2,...,m^{T_e}_{T_a})\},
\end{equation}
where $m_j^i = F(\hat{y}^i_j)$.  The mean and variance of each subset are computed to obtain
\begin{equation}
    M' = \{ (\mu_{m^1}, \sigma^2_{m^1}), ....(\mu_{m^{T_e}}, \sigma^2_{m^{T_e}}) \}.
\end{equation}

For the input image and metric, we compute the mean, $\mu_F$, as well as the aleatoric $\sigma^2_{F_a}$ and epistemic $\sigma^2_{F_e}$ uncertainties
\begin{equation}
    \mu_F = \mathbb{E}_i[\mu_i], \quad  \sigma^2_{F_a} = \mathbb{E}_i[\sigma^2_i], \quad \sigma^2_{F_e} = \mathbb{V}_i[\mu_i].
\end{equation}

Finally, both components of the clinical metric uncertainty can be combined to obtain the predictive uncertainty. 
\begin{equation}
    \sigma^2_{F} = \sigma^2_{F_a} + \sigma^2_{F_e}.
\end{equation}

The spatial uncertainty can be estimated per pixel by computing entropy of all samples in $\hat{Y}$, 
\begin{equation}
\label{eq:uncertainty_map}
    \hat{u} = H(\hat{Y}) = - \frac{1}{\log{C}} \sum^C_{c=1} \bar{y}_c \log{\bar{y}_c} \in [0,1]^{H \times W} 
\end{equation}
where $\Bar{y} = \frac{1}{T} \Sigma_{t=1}^T \hat{Y}_t \in \mathbb{R}^{C \times H \times W}$ is the mean of all samples with $T=T_a \cdot T_e$. 

\section{Experimental setup}

\begin{table*}[tp]
    \centering
    \small
    \begin{tabular*}{\textwidth}{ll  @{\extracolsep{\fill}} cccccccc}
    \toprule
    \multicolumn{2}{c}{Method} & \multicolumn{4}{c}{CAMUS} & \multicolumn{4}{c}{Private Card. US}  \\
     \cmidrule(lr){1-2} \cmidrule(lr){3-6}  \cmidrule(lr){7-10} 

        Aleatoric & Epistemic & \mcc{Dice $\uparrow$} & \mcc{Corr. $\uparrow$} & \mcc{ECE$\downarrow$} & \mcc{MI$\uparrow$}  & \mcc{Dice$\uparrow$} & \mcc{Corr.$\uparrow$} & \mcc{ECE$\downarrow$} & \mcc{MI$\uparrow$}  \\
        
    \midrule\midrule

PWA      & \multicolumn{1}{c}{-}     &.932\tiny{$\pm$.001}	      & .057\tiny{$\pm$.019}	    & .045\tiny{$\pm$.005}	    	& .135\tiny{$\pm$.006}        & .876\tiny{$\pm$0.004}	    	&	.139\tiny{$\pm$.110}          &	.282\tiny{$\pm$.009}	     & 0.127\tiny{$\pm$0.005}       \\
TTA             & \multicolumn{1}{c}{-}     &.931\tiny{$\pm$.002}	      & .252\tiny{$\pm$.022}	    & \blue{.044\tiny{$\pm$.003}}	& \blue{.209\tiny{$\pm$.004}} & .873\tiny{$\pm$0.003}	    	&	.421\tiny{$\pm$.046}          &	\red{.073\tiny{$\pm$.001}}	 & 0.199\tiny{$\pm$0.005}       \\
SSN             & \multicolumn{1}{c}{-}     &.927\tiny{$\pm$.005}	      & .172\tiny{$\pm$.060}	    & .085\tiny{$\pm$.022}	    	& .188\tiny{$\pm$.012}        & .868\tiny{$\pm$0.004}	    	&	.233\tiny{$\pm$.055}          &	.168\tiny{$\pm$.030}	     & 0.204\tiny{$\pm$0.009}       \\
CASUS           & \multicolumn{1}{c}{-}     &\red{.934\tiny{$\pm$.003}}	  & \blue{.666\tiny{$\pm$.142}} & .083\tiny{$\pm$.008}	    	& .189\tiny{$\pm$.003}        & .875\tiny{$\pm$0.005}	    	&	\red{.609\tiny{$\pm$.162}}    &	.102\tiny{$\pm$.007}	     & \blue{0.211\tiny{$\pm$0.005}}       \\
CASUS+t         & \multicolumn{1}{c}{-}     &\red{.934\tiny{$\pm$.003}}	  & \red{.674\tiny{$\pm$.138}}	& .080\tiny{$\pm$.009}	    	& .200\tiny{$\pm$.008}        & .875\tiny{$\pm$0.005}	    	&	\blue{.606\tiny{$\pm$.147}}   &	.093\tiny{$\pm$.007}	     & \red{0.222\tiny{$\pm$0.005}}  \\ \midrule
\multicolumn{1}{c}{-} & MC-D.               &.931\tiny{$\pm$.002}	      & .071\tiny{$\pm$.051}	    & \red{.042\tiny{$\pm$.005}}	& .156\tiny{$\pm$.017}        & .875\tiny{$\pm$0.002}	    	&	.138\tiny{$\pm$.039}          &	\blue{.076\tiny{$\pm$.005}}	 & 0.155\tiny{$\pm$0.012}      		\\ \midrule
PWA  & MC-D.                         &\blue{.933\tiny{$\pm$.000}}  & .065\tiny{$\pm$.050}	    & \blue{.044\tiny{$\pm$.000}}	& .131\tiny{$\pm$.012}        & \red{.880\tiny{$\pm$0.003}}		&	.018\tiny{$\pm$.061}          &	.083\tiny{$\pm$.006}	     & 0.126\tiny{$\pm$0.014}        \\
TTA  & MC-D.                                &.932\tiny{$\pm$.002}	      & .338\tiny{$\pm$.094}	    & .047\tiny{$\pm$.003}	    	& \red{.214\tiny{$\pm$.005}}  & .875\tiny{$\pm$0.002}	    	&	.472\tiny{$\pm$.045}          &	.083\tiny{$\pm$.002}	     & 0.206\tiny{$\pm$0.005}        \\
SSN & MC-D.                                 &.928\tiny{$\pm$.002}	      & .181\tiny{$\pm$.089}	    & .104\tiny{$\pm$.039}	    	& .173\tiny{$\pm$.015}        & .874\tiny{$\pm$0.004}	    	&	.240\tiny{$\pm$.094}          &	.205\tiny{$\pm$.051}	     & 0.178\tiny{$\pm$0.014}        \\
CASUS & MC-D.                               &\red{.934\tiny{$\pm$.001}}	  & .624\tiny{$\pm$.046}	    & .090\tiny{$\pm$.001}	    	& .178\tiny{$\pm$.003}        & \blue{.879\tiny{$\pm$0.001}}	&	.588\tiny{$\pm$.113}          &	.108\tiny{$\pm$.006}	     & 0.198\tiny{$\pm$0.004}      		\\
CASUS+t & MC-D.                             &\red{.934\tiny{$\pm$.001}}	  & .652\tiny{$\pm$.047}	    & .086\tiny{$\pm$.003}	    	& .188\tiny{$\pm$.002}        & \blue{.879\tiny{$\pm$0.001}}	&	.558\tiny{$\pm$.092}          &	.097\tiny{$\pm$.006}	     & 0.209\tiny{$\pm$0.004}          \\

    \bottomrule  & 
    \end{tabular*}
    \caption{Mean and standard deviation values segmentation results (5 random seeds). Best and second best results are highlighted in \red{red} and \blue{blue} respectively.}    
    \label{tab:seg_results}
\end{table*}

\subsection{Data}

\subsubsection{CAMUS} 
The CAMUS dataset, as detailed in the work by Leclerc et al.~\cite{Leclerc19}, encompasses conventional clinical examinations derived from 500 patients and acquired using a GE Vivid E95 ultrasound scanner. The imaging acquisitions were specifically tailored for the assessment of left ventricular ejection fraction. Each patient's dataset includes 2D A2C and A4C view sequences, employing a consistent acquisition protocol and exported from the EchoPAC analysis software (GE Vingmed Ultrasound, Horten, Norway).

A senior cardiologist annotated the endocardium and epicardium borders of the left ventricle, as well as the atrium, for both ED and ES images in the dataset. The dataset is partitioned into 400 training patients, 50 validation patients, and 50 testing patients. Contour points were defined by identifying the endocardium basal points, followed by determining the apex as the farthest point along the respective edges. Each contour is composed of 21 points.

\subsubsection{Private UK dataset} The dataset is derived from the EVAREST trial~\cite{woodward2022real}, a comprehensive multi-site, multi-vendor study conducted in the UK. A2C and A4C echocardiograms were acquired from patients at rest or under induced stress. The dataset includes healthy individuals and patients with conditions such as coronary artery disease and COVID.

The private UK dataset is divided into a training/validation set (80/20 split) and an independent test set sourced from different sites. It consists of 994 echocardiograms from 684 patients for the training/validation set and 368 echocardiograms from 206 patients for the independent test set. Expert annotations for the endocardium contour were performed, with a minimum of 7 points labeled based on anatomical landmarks. Additional points were added as needed to precisely delineate the contour, and the contours were subsequently resampled to include a total of 21 evenly distributed points along the contour.

\subsubsection{Private French dataset}
\added{
We assess generalisation across pathologies with a private France dataset, acquired at 
the Hospices Civils de Lyon 
(HCL), France. This dataset contains 80 sequences divided into 4 groups of 20 patients each, covering the following pathologies: hypertrophic cardiomyopathy with amyloidosis (HCM-CA), hypertrophic cardiomyopathy with hypertension (HCM-HT), hypokinetic ischemic cardiomyopathy (ICM), and hypokinetic non-ischemic cardiomyopathy (NICM). As with the other datasets, the endocardium is labeled at the ED and ES frames.}

\begin{table*}[tp]
\centering
\small
\begin{tabular*}{\textwidth}{ll  @{\extracolsep{\fill}} cccccccc}
\toprule
\multicolumn{2}{c}{Method} & \multicolumn{4}{c}{CAMUS} & \multicolumn{4}{c}{Private Card. US}  \\
\cmidrule(lr){1-2} \cmidrule(lr){3-6}  \cmidrule(lr){7-10} 

Aleatoric & Epistemic & \mcc{Area} & \mcc{FAC} &  \mcc{Volume}  & \mcc{EF} & \mcc{Area} & \mcc{FAC}  & \mcc{Volume}  & \mcc{EF}   \\

\midrule\midrule

    PWA  & \multicolumn{1}{c}{-} &	3114\tiny{$\pm$115}       &	45.7\tiny{$\pm$1.9}\small{(0.8)}          &	59.7\tiny{$\pm$2.7}       	&	21.8\tiny{$\pm$2.5}\small{(0.0)}      	& 30968\tiny{$\pm$1718}         &   366.3\tiny{$\pm$12.6}\small{(3.6)}      &	550.6\tiny{$\pm$9.3}          &	156.9\tiny{$\pm$7.0}\small{(0.8)}         \\    
    TTA  & \multicolumn{1}{c}{-}        &	2685\tiny{$\pm$166}       &	32.9\tiny{$\pm$3.9}\small{(0.8)}          &	58.4\tiny{$\pm$3.7}       	&	15.9\tiny{$\pm$3.2}\small{(0.0)}      	& 24565\tiny{$\pm$538}          &   281.2\tiny{$\pm$12.6}\small{(3.6)}      &	454.1\tiny{$\pm$17.2}         &	135.9\tiny{$\pm$5.7}\small{(0.9)}         \\    
    SSN & \multicolumn{1}{c}{-}         &	1866\tiny{$\pm$615}       &	57.4\tiny{$\pm$6.6}\small{(0.2)}          &	25.7\tiny{$\pm$4.0}       	&	18.9\tiny{$\pm$3.5}\small{(0.0)}      	& 11528\tiny{$\pm$6290}         &   312.9\tiny{$\pm$58.6}\small{(6.2)}      &	156.7\tiny{$\pm$48.6}         &	68.7\tiny{$\pm$18.5}\small{(2.5)}         \\    
    CASUS & \multicolumn{1}{c}{-}       &	539\tiny{$\pm$194}        &	18.4\tiny{$\pm$3.9}\small{(0.6)}          &	13.5\tiny{$\pm$6.9}       	&	8.4\tiny{$\pm$2.2}\small{(0.0)}       	& 10546\tiny{$\pm$1179}         &   \blue{42.0\tiny{$\pm$4.1}\small{(3.6)}} &	250.3\tiny{$\pm$34.6}         &	\red{22.1\tiny{$\pm$2.2}\small{(0.8)}}    \\    
    CASUS+t & \multicolumn{1}{c}{-}     &	609\tiny{$\pm$153}        &	\red{10.0\tiny{$\pm$2.0}\small{(0.4)}}    &	17.3\tiny{$\pm$5.2}       	& \red{6.0\tiny{$\pm$0.5}\small{(0.0)}} 	& 8680\tiny{$\pm$608}           &   49.4\tiny{$\pm$8.6}\small{(2.7)}        &	219.0\tiny{$\pm$27.0}         &	36.4\tiny{$\pm$6.1}\small{(0.0)}          \\   \midrule 
    \multicolumn{1}{c}{-} & MC-D        &	3460\tiny{$\pm$169}       &	48.6\tiny{$\pm$4.4}\small{(0.8)}          &	72.1\tiny{$\pm$7.3}       	&	27.4\tiny{$\pm$4.8}\small{(0.4)}      	& 30570\tiny{$\pm$790}          &   374.2\tiny{$\pm$12.2}\small{(3.9)}      &	579.8\tiny{$\pm$28.9}         &	182.6\tiny{$\pm$5.1}\small{(1.2)}         \\   \midrule
    PWA  & MC-D                  &	3101\tiny{$\pm$96}        &	46.6\tiny{$\pm$4.0}\small{(0.8)}          &	57.9\tiny{$\pm$2.8}       	&	23.1\tiny{$\pm$3.6}\small{(0.0)}      	& 28914\tiny{$\pm$1438}         &   351.1\tiny{$\pm$15.4}\small{(3.3)}      &	508.3\tiny{$\pm$26.9}         &	151.2\tiny{$\pm$8.1}\small{(0.5)}         \\    
    TTA  & MC-D                         &	2430\tiny{$\pm$203}       &	28.1\tiny{$\pm$3.8}\small{(1.2)}          &	48.1\tiny{$\pm$7.8}       	&	14.2\tiny{$\pm$4.0}\small{(0.4)}      	& 22757\tiny{$\pm$1153}         &   250.7\tiny{$\pm$7.6}\small{(0.0)}       &	417.7\tiny{$\pm$26.3}         &	115.0\tiny{$\pm$5.1}\small{(0.0)}         \\    
    SSN & MC-D                          &	3618\tiny{$\pm$1241}      &	87.3\tiny{$\pm$19.1}\small{(0.8)}         &	46.7\tiny{$\pm$16.9}      	&	30.3\tiny{$\pm$7.0}\small{(0.0)}      	& 25388\tiny{$\pm$8890}         &   535.6\tiny{$\pm$113.1}\small{(4.3)}     &	267.3\tiny{$\pm$174.9}        &	149.2\tiny{$\pm$37.4}\small{(1.5)}        \\    
    CASUS & MC-D                        &	\blue{499\tiny{$\pm$145}} &	27.6\tiny{$\pm$1.6}\small{(0.8)}          &	\blue{12.7\tiny{$\pm$1.9}}	&	13.6\tiny{$\pm$1.5}\small{(0.0)}      	& \blue{5365\tiny{$\pm$1177}}   &   80.8\tiny{$\pm$9.7}\small{(2.9)}        &	\blue{147.5\tiny{$\pm$37.7}}  &	34.0\tiny{$\pm$9.6}\small{(0.7)}          \\    
    CASUS+t & MC-D                      &	\red{420\tiny{$\pm$122}}  &	\blue{15.2\tiny{$\pm$2.5}\small{(0.6)}}   &	\red{11.6\tiny{$\pm$4.8}} 	&	\blue{8.4\tiny{$\pm$1.0}\small{(0.0)}}	& \red{4577\tiny{$\pm$1041}}    &   \red{31.6\tiny{$\pm$7.9}\small{(2.7)}}  &	\red{118.2\tiny{$\pm$28.4}}   &	\blue{23.8\tiny{$\pm$8.4}\small{(0.0)}}   \\    
    \bottomrule      
    \end{tabular*}
    \caption{Mean and standard deviation for uncertainty calibration error values (5 random seeds). Best and second best results are highlighted in \red{red} and \blue{blue} respectively. The average percentage of rejected samples is indicated in parentheses.}    

    \label{tab:clinical_results}
\end{table*}

\subsection{Model Configuration}

To evaluate both our uncertainty propagation framework and our contour sampling method, we implemented a variety of aleatoric uncertainty methods for segmentation. We compared our method to pixel-wise aleatoric uncertainty  (\textbf{PWA})~\cite{bayesianCV}, stochastic segmentation networks (\textbf{SSN})~\cite{stochastic_seg_net} and test-time augmentation (\textbf{TTA})~\cite{WANG2019_tta_seg}. We also evaluate our method (\textbf{CASUS}), and our method with the temporal consideration described in \Cref{sec:temporal_sampling} (\textbf{CASUS-t}). We also compare with epistemic uncertainty using Monte-Carlo Dropout (\textbf{MC-D}). Finally, we combine both aleatoric and epistemic uncertainty for all combinations of aleatoric and epistemic methods.
For all methods, we use a 2D U-Net model defined in \cite{camus_nnUnet} and resized images to 256x256 pixels. All methods were trained with a batch size of 32 for a maximum of 1000 epochs with early stopping after 100 epochs without validation loss improvement. The Adam optimizer \cite{adam} was used for all methods except for the stochastic segmentation network, which was only trainable with RSM-prop which is in the original code provided by the authors. We trained using data augmentation, which included random rotations and translations, random brightness and contrast changes, and finally, random Gamma corrections. We used the same data augmentation for the test-time augmentations. 

We generated samples with $T_a=25$ and $T_e=10$, except for the MC-Dropout only method where $T_e=25$. We applied post processing to both the mean prediction and the Monte-Carlo samples for all segmentation based methods to remove holes and select only the largest predicted structure. 

\subsection{Evaluation Metrics}


\subsubsection{Segmentation metrics}

For the segmentation metrics, we consider the output segmentation map, as well as the uncertainty map, predicted using \Cref{eq:uncertainty_map}. We used the \textbf{Dice} coefficient to assess the segmentation quality. To assess the quality of the uncertainty map we consider the image-level uncertainty estimate by computing the mean of the uncertainty map weighted by the size of the prediction and pixel-level uncertainty, where the raw pixel uncertainty values are used. The metric used to assess the image level uncertainty is the negative correlation (\textbf{Corr.}) between image-level uncertainty and the image Dice. To evaluate the pixel-wise calibration of the uncertainty maps the expected calibration error (\textbf{ECE}) is computed as follows~\cite{ece}: 
\begin{equation}
\text{ECE} = \sum_{m=1}^M \frac{|B_m|}{N_{test}} | \text{acc}(B_m) - \text{conf}(B_m)|
\end{equation}
where $\text{acc}(B_m)$ is the accuracy and $\text{conf}(B_m)$ is the average confidence for a batch $B_m$. As the uncertainty map values are bounded between 0 and 1, we define the confidence $c$ for each pixel as $c = 1 - u$ where $u$ is the pixel uncertainty. 

Finally, to assess if the uncertainty is predicted in the correct area, the mutual information (\textbf{MI}) between the uncertainty map and the pixel-wise error map computed between the mean prediction and ground truth segmentation map is computed.

\subsubsection{Clinical metrics}

For each segmentation output, we compute 4 clinical metrics: the left ventricular area (\textbf{Area}), the left-ventricular fractional area change (\textbf{FAC}), the left ventricular volume ES and ED using the 2CH and 4CH views (\textbf{Volume)}, and the ejection fraction (\textbf{EF}). \added{We compute the reference volumes using the Simpson biplane method using the code released with the CAMUS dataset to stay consistent with previous works. Although widely used, the reader shall note that the Simpson biplane method can be subject to geometric inaccuracies arising from LV foreshortening, basal slanting, and assumptions regarding the orientation and ellipticity of LV cross-sections~\cite{kim2023beyond}.  As such, volume results should be interpreted with care.}

To assess the uncertainty estimates, we use the uncertainty calibration error (\textbf{UCE}) \cite{pmlr-v121-laves20a_UCE}, defined by
\begin{equation}
\text{UCE} = \sum_{m=1}^M \frac{|B_m|}{N_{test}} | \text{err}(B_m) - \text{uncert}(B_m)|,
\end{equation}
where $\text{err}(B_m)$ and $\text{uncert}(B_m)$ are the average error and uncertainty for a batch $B_m$. We define the bins to contain an equal number of samples, as is used when computing adaptive calibration error~\cite{Nixon_2019_CVPR_Workshops_ACE}, because the uncertainty values can have very large ranges which makes equal bin ranges likely to contain few or no samples. 

In certain cases, model predictions or samples generated by various methods yield values that are clinically irrelevant. For example, due to the independent estimation of ES and ED samples, metrics such as FAC or ejection EF may result in negative values, which are not meaningful in a clinical context. Such predictions would be discarded in a clinical setting due to their irrelevance, irrespective of associated uncertainty. To ensure that these clinically irrelevant samples do not bias the uncertainty calibration error calculation, we exclude them based on the following criteria:

\begin{enumerate}
    \item Reject predictions with negative values.
    \item Discard Monte Carlo samples with negative values.
    \item Reject a prediction if more than 50\% of its Monte-Carlo samples have been discarded.
\end{enumerate}

We report the percentage of rejected samples in the test set for both FAC and EF across all evaluated methods.

\begin{figure*}[tp]
  \centering
\begingroup
    \captionsetup[subfigure]{labelformat=empty,labelsep=none}
    \subfloat[]{
      \includegraphics[width=\textwidth]{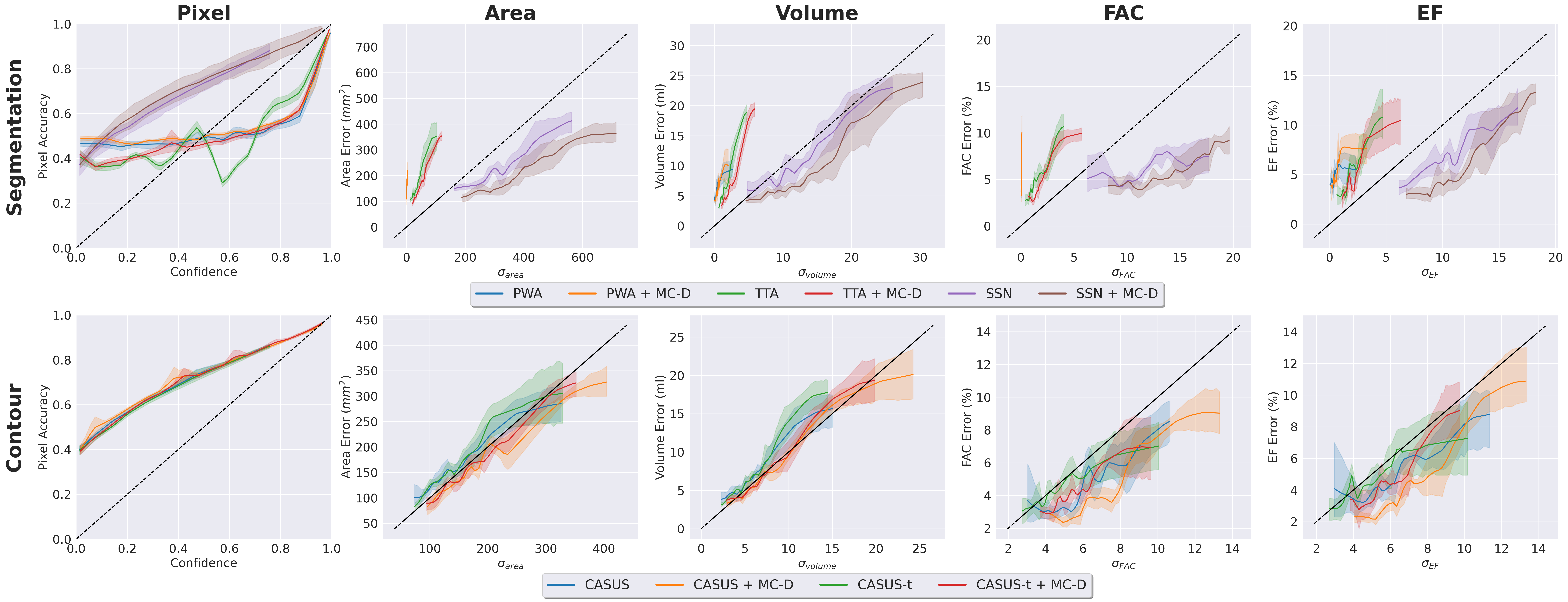}
    }\hfill
    \subfloat[]{
      \includegraphics[width=\textwidth]{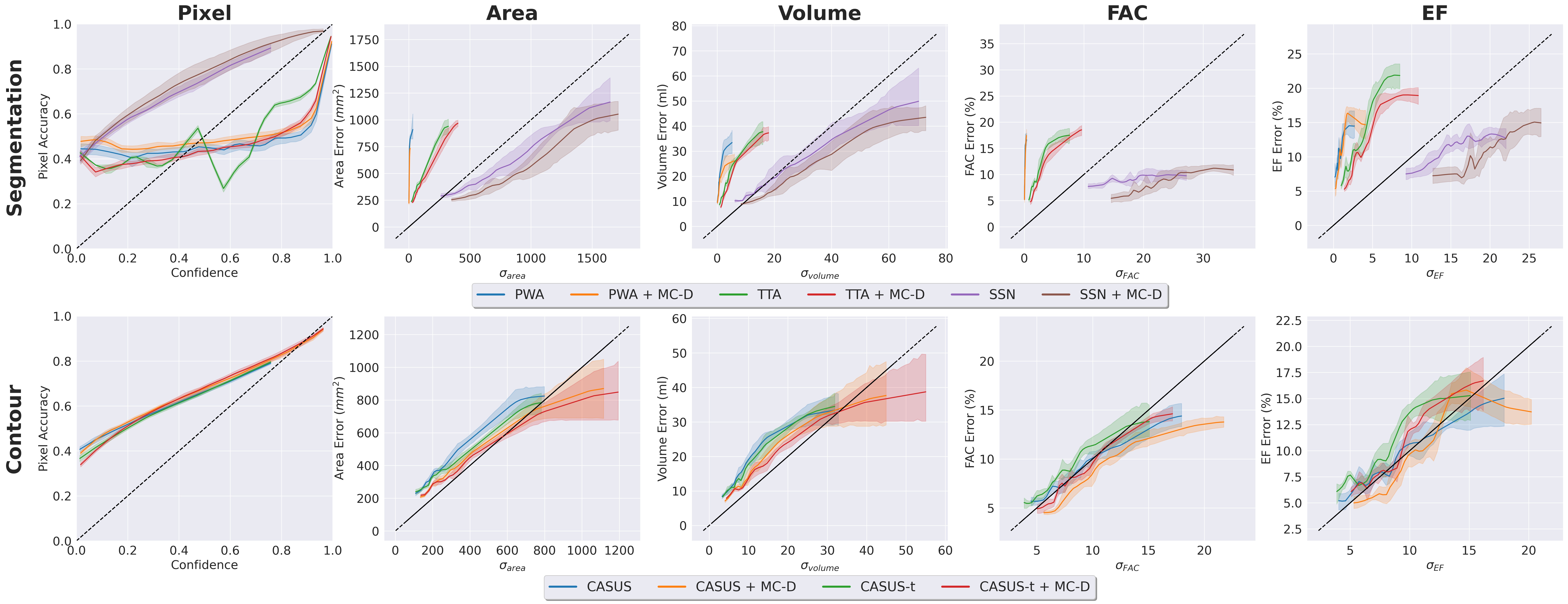}
    }
  \endgroup
    \caption{Reliability diagrams for the [TOP] CAMUS and [BOTTOM] private UK datasets. Each row shows the calibration diagram for each aleatoric method with and without MC Dropout for each metric by column. Dashed lines indicate perfect calibration.}
\label{fig:calibration}
\end{figure*}


\section{Results}

\subsection{Segmentation Metrics}

Segmentation results are shown in \Cref{tab:seg_results}. Samples showing the error and uncertainty maps used for the different metric calculations are shown in \Cref{fig:samples}.  All methods exhibit comparable Dice scores across both datasets, with the added complexity of the private UK dataset leading to lower scores across methods. Since the overarching objective of our method is to propagate uncertainty from the input images to the derived metrics, this behavior is desired as it shows that our segmentation results are on par with other state-of-the-art methods. Indeed, all methods achieve adequate performance on the segmentation task, and in turn, no method had an advantage for the uncertainty metrics. For uncertainty estimation, all contouring methods outperform segmentation approaches in terms of correlation between the Dice and uncertainty map mean; the highest correlation being CASUS-t. 

Expected calibration error results indicate that MC Dropout and test-time augmentation have the best calibration performance. However, the pixel confidence reliability diagrams~\cite{reliability_diagram} in \Cref{fig:calibration}(col. 1) reveal that none of the segmentation methods achieve ideal calibration, as no method perfectly follows the diagonal calibration line. Most segmentation methods' lower ECE values are influenced by high-confidence bins containing the highest number of pixels, as shown in \Cref{fig:samples}’s individual reliability diagrams. This can be explained by the behavior observed in \Cref{fig:samples}, showing that uncertainty is predicted just around the border of the prediction for methods like PWA and TTA. Conversely, CASUS and SSN generate more variability for all images, leading to over-estimating the uncertainty on the full dataset calibration, but resulting in better calibration for images with higher error. 

Lastly, the uncertainty-error mutual information metric in \Cref{tab:seg_results} highlights that test-time augmentation and CASUS are the most effective in predicting uncertainty regions accurately. TTA performs better on the CAMUS dataset, where errors are smaller and predicted uncertainty aligns with errors. Conversely, on the Private UK dataset, where errors are more substantial, CASUS yields higher mutual information values, suggesting it captures larger errors more effectively.

Overall, adding epistemic uncertainty to aleatoric uncertainty has a minimal impact on segmentation metrics. This may be due to aleatoric uncertainty accounting for a much larger proportion of the full predictive uncertainty, and that there is no specific distinction between aleatoric and epistemic uncertainty when computing the uncertainty map.

\subsection{Clinical Metrics}


The clinical uncertainty metric results for both the CAMUS and private UK datasets are summarized in \Cref{tab:clinical_results}, with reliability diagrams displayed in \Cref{fig:calibration}. Pixel-wise aleatoric uncertainty performs the worst, as reflected in both the metrics and reliability diagrams. This poor performance is attributed to limited variability in the generated samples, as seen in \Cref{fig:samples}, resulting in highly over-confident estimates. Although TTA introduces slightly more variability, it remains insufficient to achieve reliable calibration for clinical uncertainty. 

The limited variability of both methods, as seen in \Cref{fig:samples}, results in negligible uncertainty in the estimated clinical metrics and thus a large uncertainty calibration error. The reliability diagrams in \Cref{fig:calibration} further illustrate this issue: although PWA and TTA cover error ranges comparable to those of other approaches (y-axis), they predict very little uncertainty (x-axis), leading to calibration curves that diverge markedly from the ideal diagonal.

In contrast, the stochastic segmentation networks (SSN) and our proposed method, CASUS, produce more diverse samples (\Cref{fig:samples}) and therefore yield meaningful uncertainty estimates when appropriate (\Cref{fig:calibration}). SSN achieves this by modeling a richer distribution that captures pixel-wise covariance, allowing for more expressive contour predictions. We attribute the superior performance of CASUS to its explicit modeling of localization uncertainty, which, when combined with the shape-uncertainty model, results in the most reliable uncertainty estimates.

Stochastic segmentation networks and CASUS variants perform similarly on area and volume metrics, with CASUS showing a slight advantage. CASUS-t, however, significantly outperforms SSN in FAC and EF metrics, as SSN lacks the ability to model temporal coherence between frames.

In contrast to segmentation metrics, modeling epistemic uncertainty substantially improves calibration for clinical metrics, especially on the private UK dataset, where there is a larger domain shift between training and test data. When only predicting the aleatoric component of the uncertainty, CASUS shows slightly over-confident uncertainty estimates. With the addition of epistemic uncertainty, provided by MC Dropout, the predictions are almost perfectly calibrated. 

CASUS+t method combined with  MC-Dropout ranks as the top or second-best method across all metrics in both datasets, emphasizing the benefits of temporally consistent sampling and the inclusion of epistemic uncertainty.

\subsection{Generalization \added{to cardiac shapes and sizes}}

\begin{table}[]
\begin{tabular}{llccc}
\toprule
\multirow{2}{*}{Set}   & \multirow{2}{*}{Metric} & \multicolumn{3}{c}{Patient type}                                \\
\cmidrule{3-5}
                       &                         & All               & EF $<$ 40\%       & EF $\ge$ 40\%      \\ 
\midrule
\multirow{4}{*}{Train} & Num. patients           & 400               & 124               & 276              \\
                       & Vol. ES (ml)            & 58~[12--315] 	& 86~[28--315] 	& 45~[12--118] 	 \\
                       & Vol. ED (ml)            & 100~[31--341] 	& 123~[38--341] & 89~[31--208] 	 \\
                       & EF (\%)                 & 44~[6--82] 		& 31~[6--39] 	& 51~[40--82] 	 \\  
\midrule
\multirow{4}{*}{Test}  & Num. patients           & 50               & 15                & 35               \\
                       & Vol. ES (ml)            & 56~[12--187] 	& 85~[38--187] 	& 42~[12--93]  \\
                       & Vol. ED (ml)            & 97~[46--266] 	& 121~[52--266] & 86~[46--171] \\
                       & EF (\%)                 & 45~[5--73] 		& 30~[5--39] 	& 51~[41--73] \\
\bottomrule
\end{tabular}
\caption{Distribution of volume and ejection fraction values for the CAMUS training and testing sets for unhealthy patients (EF$<$40\%) and patients with an EF$>$40\%. The values represented are the mean, as well as the minimum and maximum.}    
\label{tab:stats}
\end{table}

\begin{table}[tp]
\centering
\footnotesize
\setlength{\tabcolsep}{3pt}
\begin{tabular}{@{}llcc@{}}
\toprule
Method & Group & \mcc{Error (ml / \%) $\downarrow$} & \mcc{UCE (ml / \%) $\downarrow$} \\
\midrule\midrule
\multirow{2}{*}{CASUS+t MC-D} & 12--69 ml  & 5.36{\tiny$\pm$0.48} / 13.4{\tiny$\pm$1.5} & 8.07{\tiny$\pm$1.18} / 22.2{\tiny$\pm$2.7} \\
 & 70--266 ml & 8.99{\tiny$\pm$0.80} / 8.1{\tiny$\pm$0.7} & 11.28{\tiny$\pm$2.62} / 12.1{\tiny$\pm$3.5} \\
\midrule
\multirow{2}{*}{SSN} & 12--69 ml & 8.74{\tiny$\pm$2.27} / 22.4{\tiny$\pm$5.2} & 13.38{\tiny$\pm$3.50} / 43.0{\tiny$\pm$6.5} \\
 & 70--266 ml   & 11.69{\tiny$\pm$0.92} / 10.8{\tiny$\pm$1.5} & 24.19{\tiny$\pm$6.78} / 27.6{\tiny$\pm$5.9} \\
\bottomrule
\end{tabular}
\caption{\added{Mean and standard deviation of dice, volume error, and volume uncertainty calibration error (5 random seeds) for test-set patients, grouped by ground-truth volume. We report both absolute and relative error to allow comparison between groups with different magnitudes.}}
\label{tab:volume_calibration_groups_merged}
\end{table}

\begin{table*}[tp]
\centering
\small
\begin{tabular*}{\textwidth}{ccc  cccccc}
\toprule
\multicolumn{3}{c}{Method} & \multicolumn{3}{c}{CAMUS (EF $< 40\%$)} & \multicolumn{3}{c}{CAMUS (EF $\ge 40\%$)}  \\
\cmidrule(lr){1-3} \cmidrule(lr){4-6}  \cmidrule(lr){7-9} 

\mcc{Aleatoric} & \mcc{Epistemic} & \mcc{Data} & \mcc{DICE $\uparrow$} &  \mcc{Volume UCE $\downarrow$}  & \mcc{EF UCE $\downarrow$} & \mcc{DICE $\uparrow$} &  \mcc{Volume UCE $\downarrow$}  & \mcc{EF UCE $\downarrow$}   \\

\midrule\midrule
SSN                 & \multicolumn{1}{c}{-}     & ALL (N=1600)                      &	.933\tiny{$\pm$.004}	&	{16.5}\tiny{$\pm$4.9}		&	8.2\tiny{$\pm$2.8}	    & .920\tiny{$\pm$.009}	&	25.5\tiny{$\pm$10.9}		&	13.7\tiny{$\pm$2.9}		  \\
SSN                 & \multicolumn{1}{c}{-}     & EF $\ge40\%$ (N=1104)               &	.930\tiny{$\pm$.007}	&	{22.7}\tiny{$\pm$11.6}		&	11.9\tiny{$\pm$3.8}	    & .915\tiny{$\pm$.005}	&	47.7\tiny{$\pm$39.9}		&	20.8\tiny{$\pm$12.0}	  \\
\midrule
CASUS+t             & MC-D                      & ALL (N=1600)                      &	.939\tiny{$\pm$.004}	&	{10.4}\tiny{$\pm$1.7}		&	4.4\tiny{$\pm$0.9}	    & .933\tiny{$\pm$.003}	&	10.1\tiny{$\pm$2.3}			&	6.9\tiny{$\pm$2.2}		  \\
CASUS+t             & MC-D                      & EF $\ge40\%$ (N=1104)               &	.937\tiny{$\pm$.003}	&	{10.7}\tiny{$\pm$4.2}		&	4.4\tiny{$\pm$0.8}	    & .931\tiny{$\pm$.003}	&	10.6\tiny{$\pm$1.2}			&	6.5\tiny{$\pm$1.7}		  \\

\bottomrule      
\end{tabular*}
\caption{Mean and standard deviation for dice and uncertainty calibration error values (5 random seeds) on patients with ejection fraction above and below 40\%.}    
\label{tab:pathologies_resut}
\end{table*}

\begin{figure*}[tp]
  \centering
  \subfloat[\added{Calibration curves for absolute UCE (top) and relative UCE (bottom) across the two volume groups in the CAMUS test set.}\label{fig:vol_cal}]{
    \includegraphics[height=6cm]{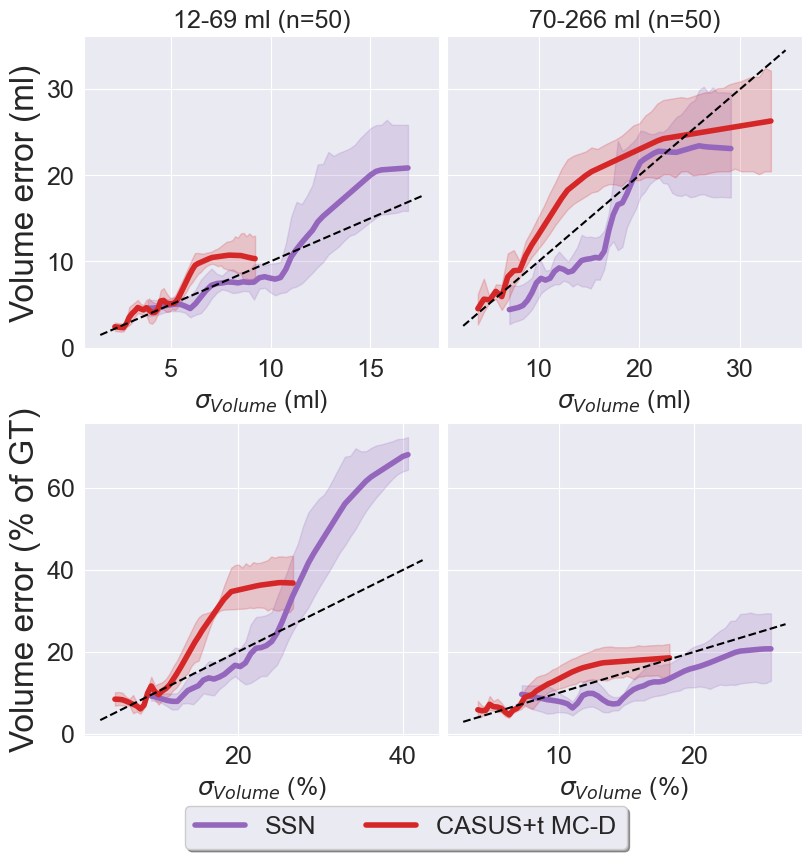}
  }
  \hspace{0.02\textwidth}
  \subfloat[\added{Calibration curves for volume (top) and ejection fraction (bottom) across the four pathologies in the private French dataset.}\label{fig:orchid_cal}]{
    \includegraphics[height=6cm]{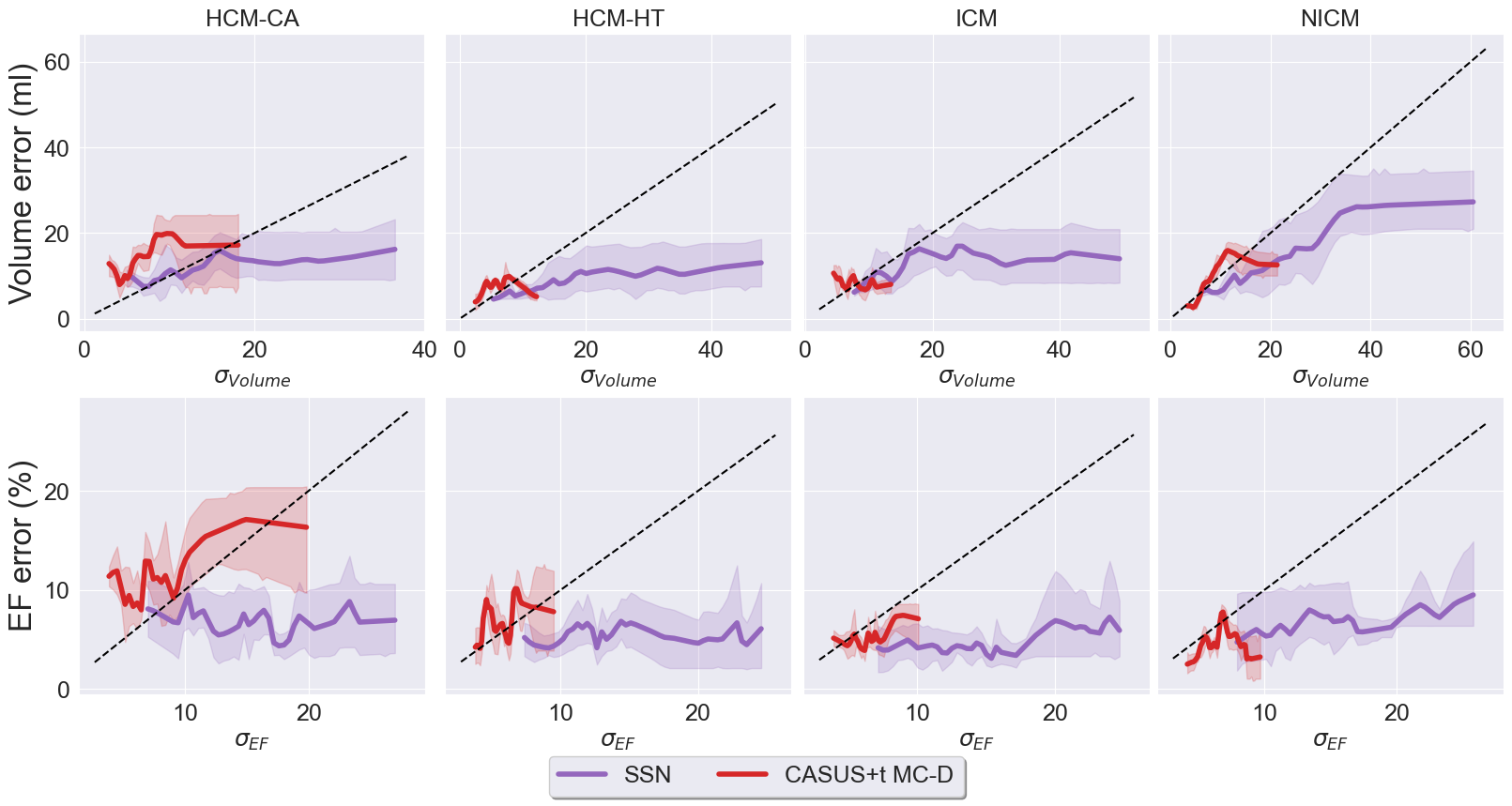}
  }
  \caption{\added{Reliability diagrams for stochastic segmentation network (SSN) and our method (CASUS+t MC-D).}}

  \label{fig:sampling_method}
\end{figure*}

\added{To assess the generalization of our method, we analyse its behavior on data subsets based on volume and ejection fraction. First, we analyse the behavior of our best-performing method CASUS-t + MC-Dropout and the best-performing segmentation method: Stochastic Segmentation Networks (SSN), on small and large volumes. We split the test set views into two groups of 50 based on volume. We report absolute error and UCE for the small and large volumes in ~\Cref{tab:volume_calibration_groups_merged}. As we compare groups where the values are of a different scale, we also report relative error and relative UCE where both the error and uncertainty are expressed as a percentage of the ground truth. Calibration curves are shown in ~\Cref{fig:vol_cal}. Relative error and relative UCE suggest that generalization is slightly more challenging for the smallest left ventricular volumes for both CASUS-t + MC-D and SSN. However, CASUS-t + MC-D consistently achieves lower errors and UCE, while also exhibiting a smaller performance gap between small and large volumes.
}

\begin{figure}[t]
    \centering

    \subfloat[\label{fig:small_lv1}]{
        \includegraphics[width=0.47\columnwidth]{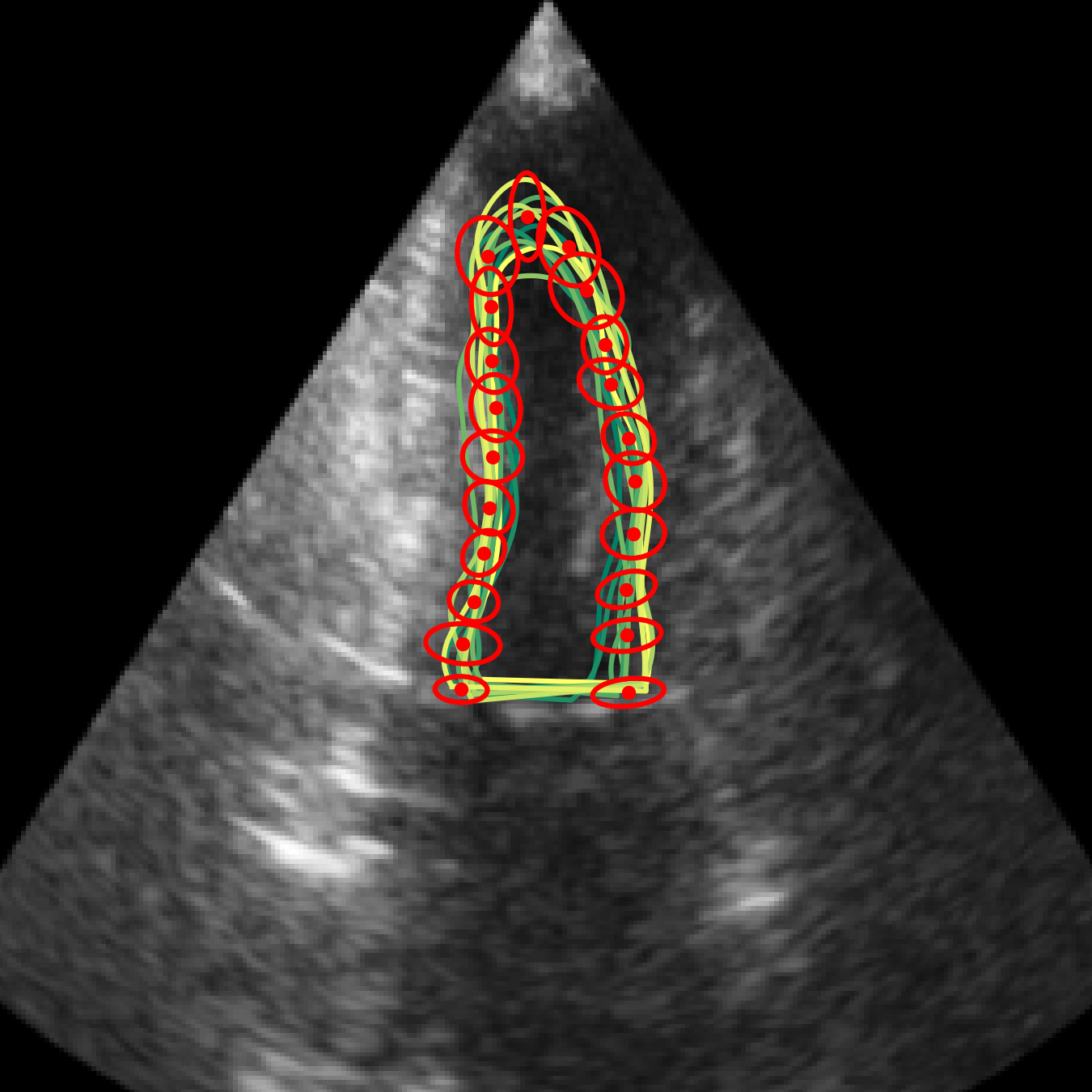}
    }
    \hfill
    \subfloat[\label{fig:small_lv2}]{
        \includegraphics[width=0.47\columnwidth]{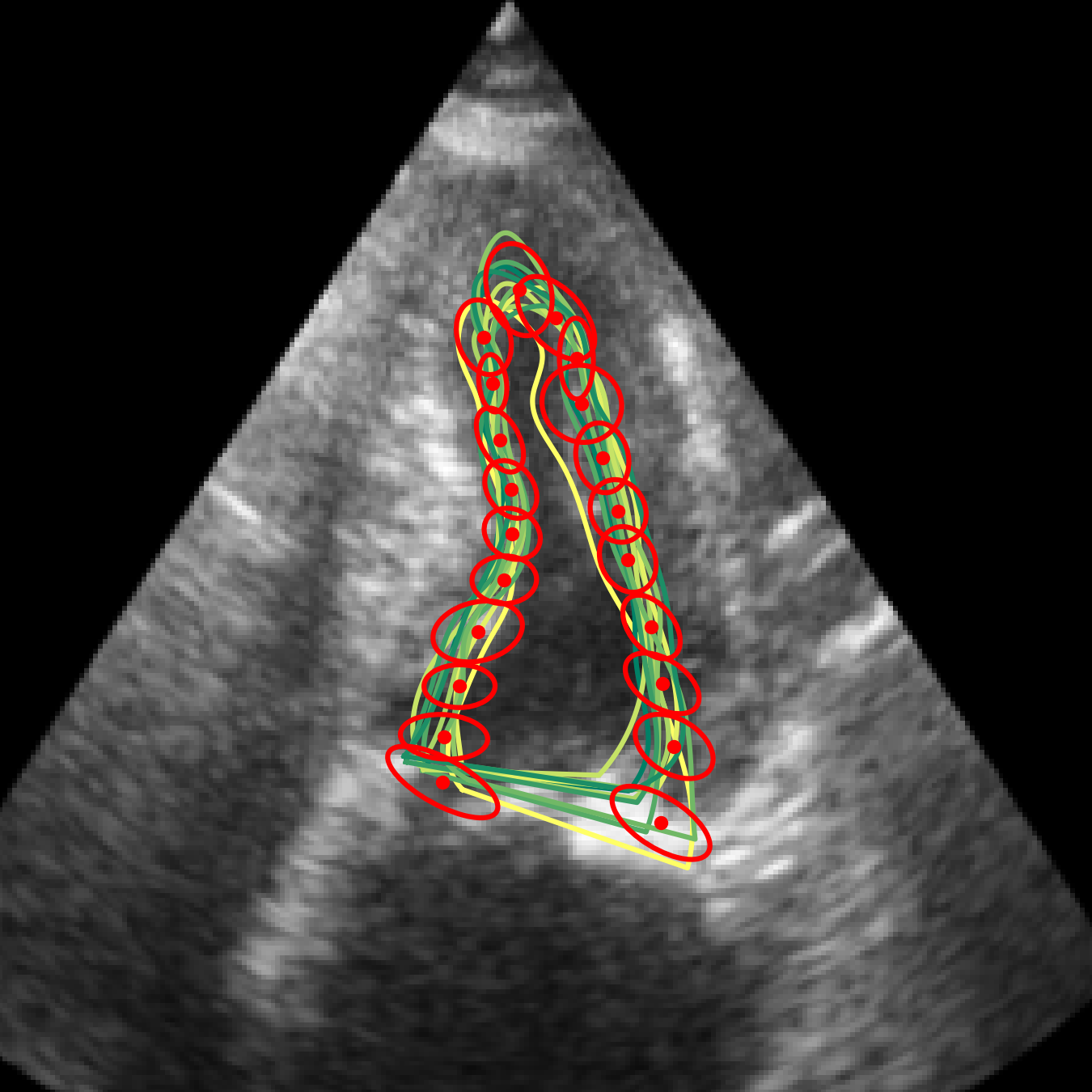}
    }


    \subfloat[\label{fig:septal_bulge}]{
        \includegraphics[width=0.98\columnwidth]{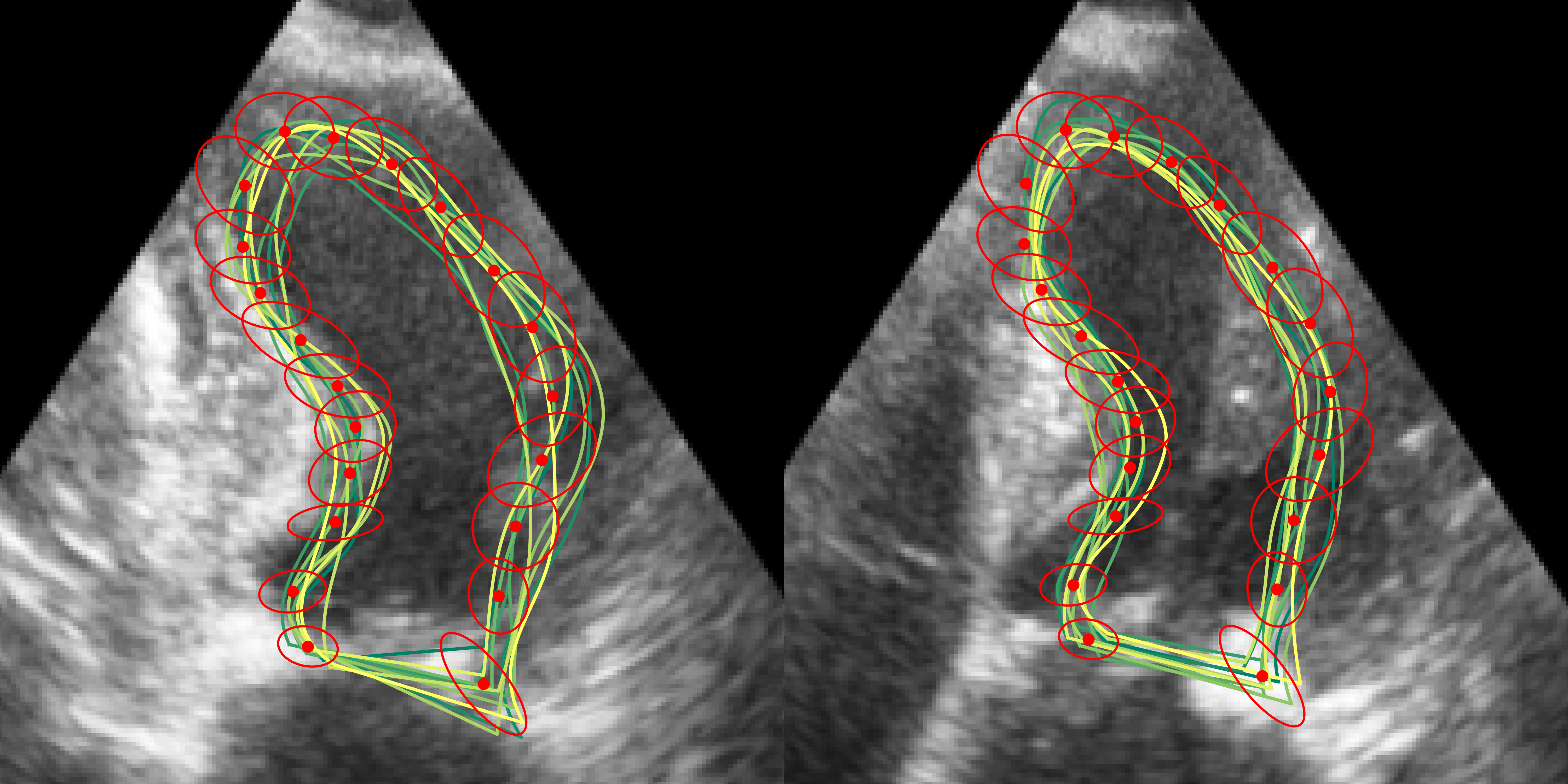}
    }

    \caption{\added{Predictive distributions and sampled contours produced by the CASUS-t for (a,b) the two samples with the smallest ES volumes (15 ml and 12 ml respectively) and} (c) the two-chamber view (left) and the four-chamber view (right) of a patient with a septal bulge \added{(EF 73\%)}. }
    \label{fig:combined}
\end{figure}

\added{To further illustrate model behavior on the smallest ventricles, we show qualitative segmentation results on the two smallest volumes of the test set in \Cref{fig:small_lv1} and \Cref{fig:small_lv2}. Despite the reduced anatomical size, the predicted contours remain anatomically plausible and close to the ground truth, though associated with relatively high uncertainty.  
Additionally,} a qualitative example of the CASUS method's performance is displayed in \Cref{fig:septal_bulge}. It shows the output of the CASUS method on two images from a patient exhibiting a septal bulge. It can be observed that the predictive uncertainty distributions in red follow the contour of the left ventricle and that the sampled contours also follow plausible left-ventricular shapes with respect to the image. This indicates that both the predictive model and the shape model are capable of generalizing to less common extreme shapes.

\replaced{We also evaluate the ability of both methods to transfer from healthy to unhealthy patients determined by ejection fraction. We split the CAMUS dataset (train and test set) into two groups.} {To assess the generalization of our method, we evaluated the behavior of the methods on two groups of patients in the CAMUS dataset based on the ejection fraction.} Group 1 contains patients with an EF lower than 40\%, meaning unhealthy patients, and group 2 contains patients with an EF greater or equal to 40\%. The statistics of the different groups of patients are given in \Cref{tab:stats} for both the training and testing sets.

\replaced{Again, we use CASUS-t + MC-Dropout and SSN.}{We selected our best-performing method CASUS-t + MC-Dropout, and the best performing segmentation method: Stochastic Segmentation Networks (SSN)}. First, both methods are trained on the full CAMUS training set and are tested separately on the two patient groups. Second, both methods are trained using only group 2. The results of this experiment are provided in \Cref{tab:pathologies_resut}. CASUS+t shows no observable performance difference between the two groups when trained on the entire dataset, and no performance degradation is observed when training is restricted to group 2. In contrast, SSN undergoes a significant drop in performance when trained solely on patients from group 2.

\begin{figure*}[tp]
  \centering
  \subfloat[\label{fig:ef_good}]{
    \includegraphics[width=0.48\textwidth]{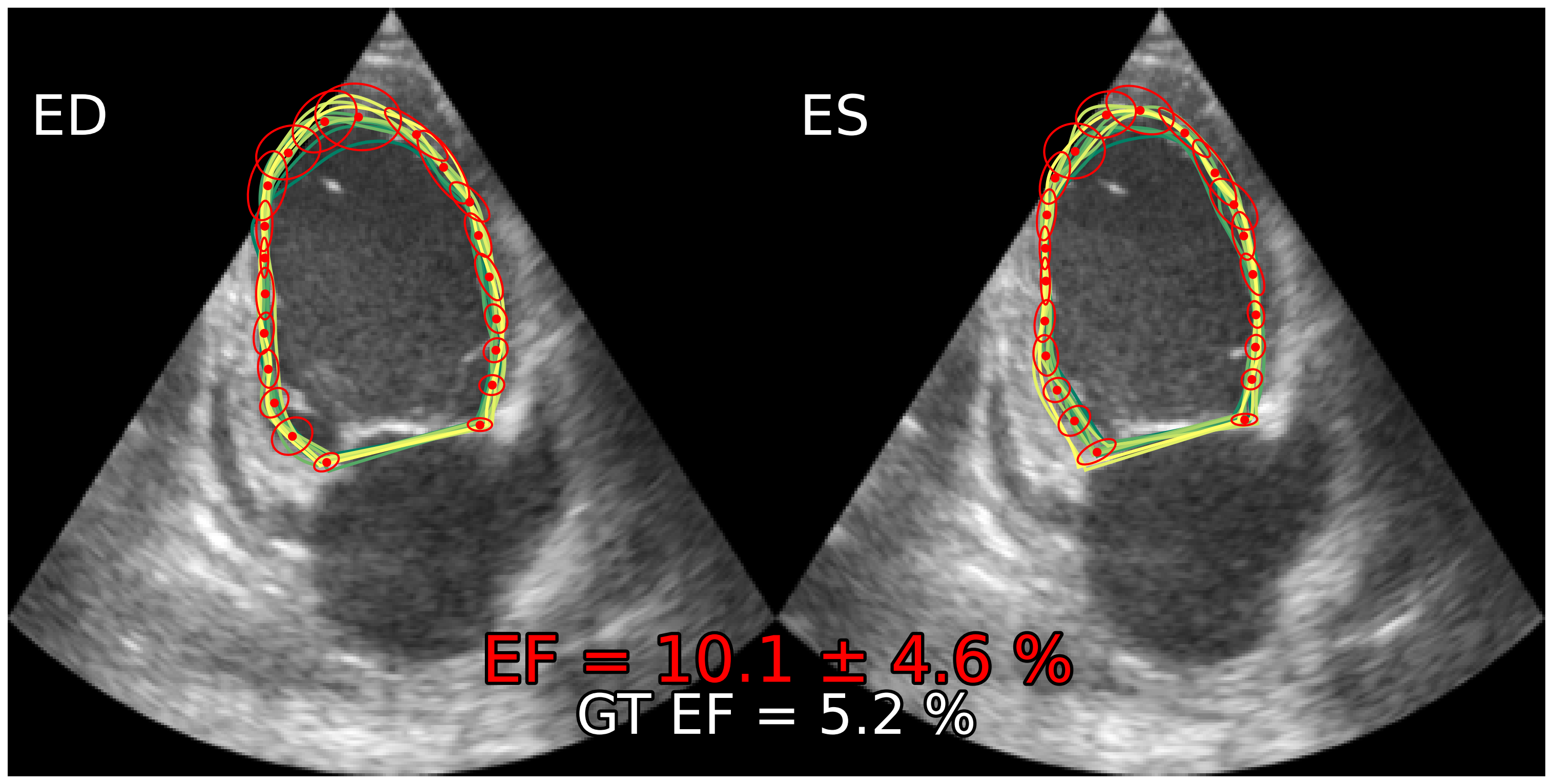}
  }\hfill
  \subfloat[\label{fig:ef_small}]{
    \includegraphics[width=0.48\textwidth]{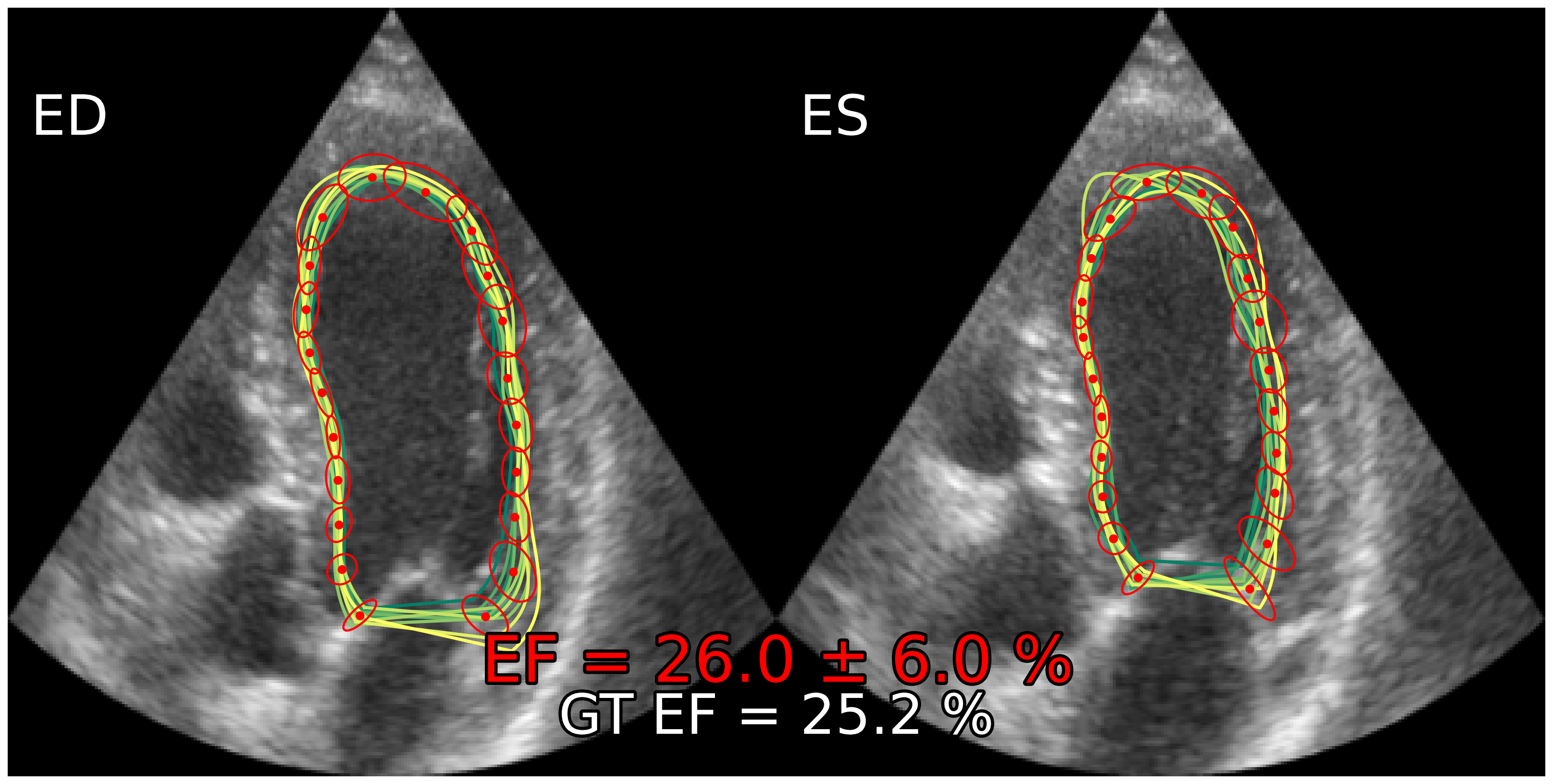}
  }


  \subfloat[\label{fig:ef_noise}]{
    \includegraphics[width=0.48\textwidth]{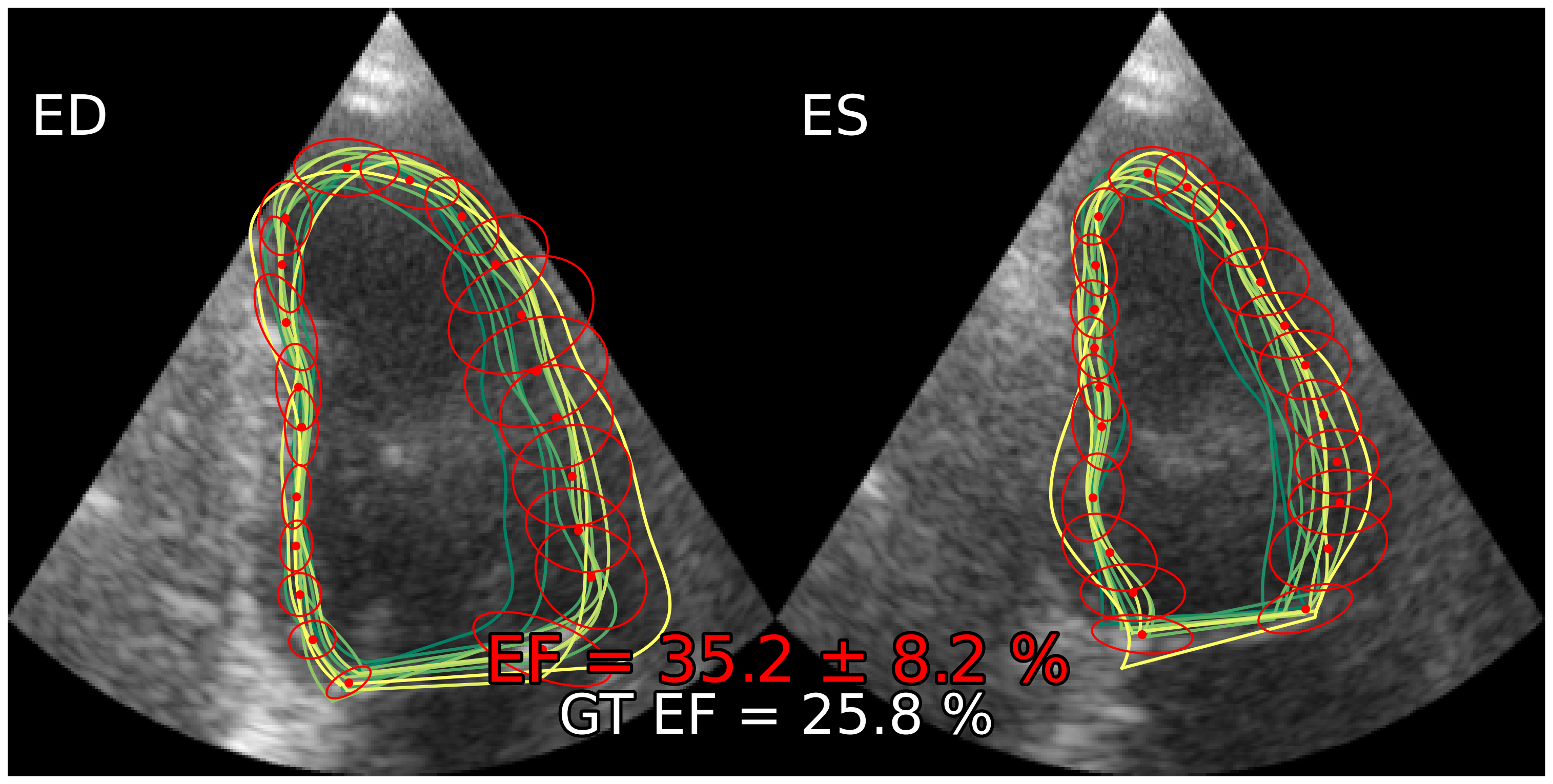}
  }\hfill
  \subfloat[\label{fig:ef_high}]{
    \includegraphics[width=0.48\textwidth]{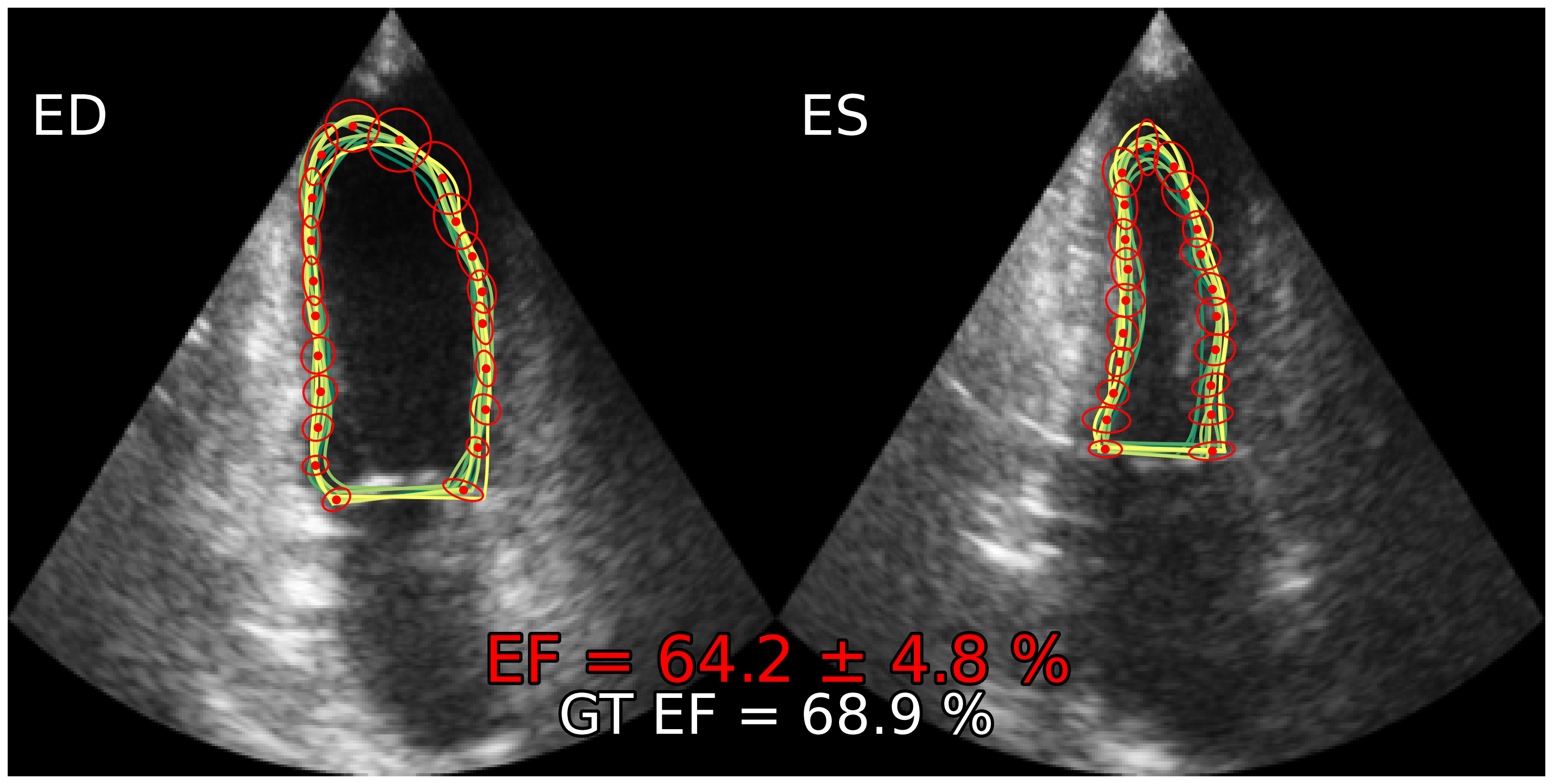}
  }

  \caption{\added{Samples from the CAMUS test set. (a) Sample with the lowest EF. (b) Sample with low EF and good image quality. (c) Sample with low EF and low contrast on the lateral wall. (c) Sample with low EF and poor image quality (d) Sample with high EF.}}
  \label{fig:ef_samples}
\end{figure*}

\Cref{fig:ef_samples} provides qualitative examples of the lowest and highest EF samples of the test set, including cases combining low EF with reduced image quality (\Cref{fig:ef_noise}). In cases with very small EF ($\leq$30\%), our method can overestimate EF relative to the ground truth. We note that this overestimation is accompanied by higher predicted uncertainty in the case with the poorest image quality (\Cref{fig:ef_noise}, ±9.4, the highest among the four examples shown), suggesting that uncertainty estimation can reflect reduced reliability in this specific case.

These findings emphasize the generalization capability of the proposed approach, as even when trained exclusively on patients with EF$\geq$40\%, the model maintains robust performance on unhealthy patients who may exhibit pathological shapes. \added{However, as seen in the qualitative examples, very small EF values can be overestimated, indicating that generalization ability can still be improved.}



\subsection{\added{Generalization to pathologies}}

\added{We assess generalisation across pathologies with the third private French dataset. As before, we compare our method to SSN. We use the models trained on the CAMUS dataset and run inference on all sequences. \Cref{tab:pathology_stats} reports the distribution of volume and ejection fraction for each group. \Cref{tab:pathology_res} reports the Dice scores and the UCE values for volume and ejection fraction for CASUS-t + MC-D and SSN. Both methods perform equally well in terms of Dice, with values almost identical to those on the training dataset. The only exception is the HCM-CA group, which shows lower performance for both methods; as this pathology is rare and possibly absent from the CAMUS dataset, the altered appearance caused by amyloidosis may degrade model performance. Calibration performance for both volume and ejection fraction, as measured by UCE, follows similar trends: it is comparable to the CAMUS dataset for all groups except HCM-CA, which again shows a drop in performance.}

\added{Further analysis of calibration is shown in the reliability diagrams in \Cref{fig:orchid_cal}. As for the previous datasets, SSN tends to be underconfident, predicting too much uncertainty relative to the observed error. Our method, by contrast, stays closer to the ideal calibration line, except for the HCM-CA group, where the increase in UCE can be attributed to overconfidence, as shown by the calibration curve lying above the diagonal of perfect calibration.}

\begin{table}[tp]
\centering
\setlength{\tabcolsep}{4pt}
\begin{tabular}{l cccc}
\toprule
 & HCM-CA & HCM-HT & ICM & NICM \\
\midrule
Num. patients & 20 & 20 & 20 & 20  \\
Vol. ES (ml)  & 54~[33--93]  & 61~[18--109]  & 87~[45--147]  & 87~[40--161] \\
Vol. ED (ml)  & 84~[51--137] & 102~[38--194] & 118~[66--175] & 116~[58--207] \\
EF (\%)       & 35~[14--52]  & 41~[19--56]   & 26~[16--40]   & 26~[14--37]  \\
\bottomrule
\end{tabular}
\caption{\added{Cohort characteristics across pathology
subsets on the private French dataset. The values represented are the mean, as well as the minimum and maximum.}}
\label{tab:pathology_stats}
\end{table}

\begin{table}[tp]
\centering
\setlength{\tabcolsep}{4pt}
\begin{tabular}{ll cccc}
\toprule
 Metric & Method & HCM-CA & HCM-HT & ICM & NICM \\
\cmidrule(lr){1-2} \cmidrule(lr){3-6}
\multirow{2}{*}{Dice $\uparrow$}
      & SSN   & .904\tiny{$\pm$.006} & .930\tiny{$\pm$.003} & .933\tiny{$\pm$.005} & .932\tiny{$\pm$.004} \\
      & CASUS-t MC-D & .898\tiny{$\pm$.002} & .929\tiny{$\pm$.002} & .940\tiny{$\pm$.002} & .940\tiny{$\pm$.004} \\
\midrule
\multirow{2}{*}{Vol. UCE $\downarrow$}
      & SSN   & 26.9\tiny{$\pm$8.7} & 37.7\tiny{$\pm$22.5} & 39.0\tiny{$\pm$25.8} & 40.6\tiny{$\pm$22.0}  \\
      & CASUS-t MC-D & 27.9\tiny{$\pm$2.4} & 11.8\tiny{$\pm$0.6 } & 11.3\tiny{$\pm$1.4} & 9.4\tiny{$\pm$0.6}   \\
\midrule
\multirow{2}{*}{EF UCE $\downarrow$}
      & SSN   & 18.9\tiny{$\pm$4.0} & 16.5\tiny{$\pm$3.7} & 18.3\tiny{$\pm$1.8} & 17.6\tiny{$\pm$2.8}  \\
      & CASUS-t MC-D & 9.6\tiny{$\pm$0.7}  & 5.7\tiny{$\pm$0.5}  & 4.1\tiny{$\pm$0.6}  & 4.9\tiny{$\pm$0.7}  \\
\bottomrule

\end{tabular}
\caption{\added{Performance of SSN and CASUS across pathology
subsets on the private French dataset.}}
\label{tab:pathology_res}
\end{table}

\begin{figure*}[tp]
  \centering
  \subfloat[\label{fig:orchid1}]{
    \includegraphics[width=0.23\textwidth]{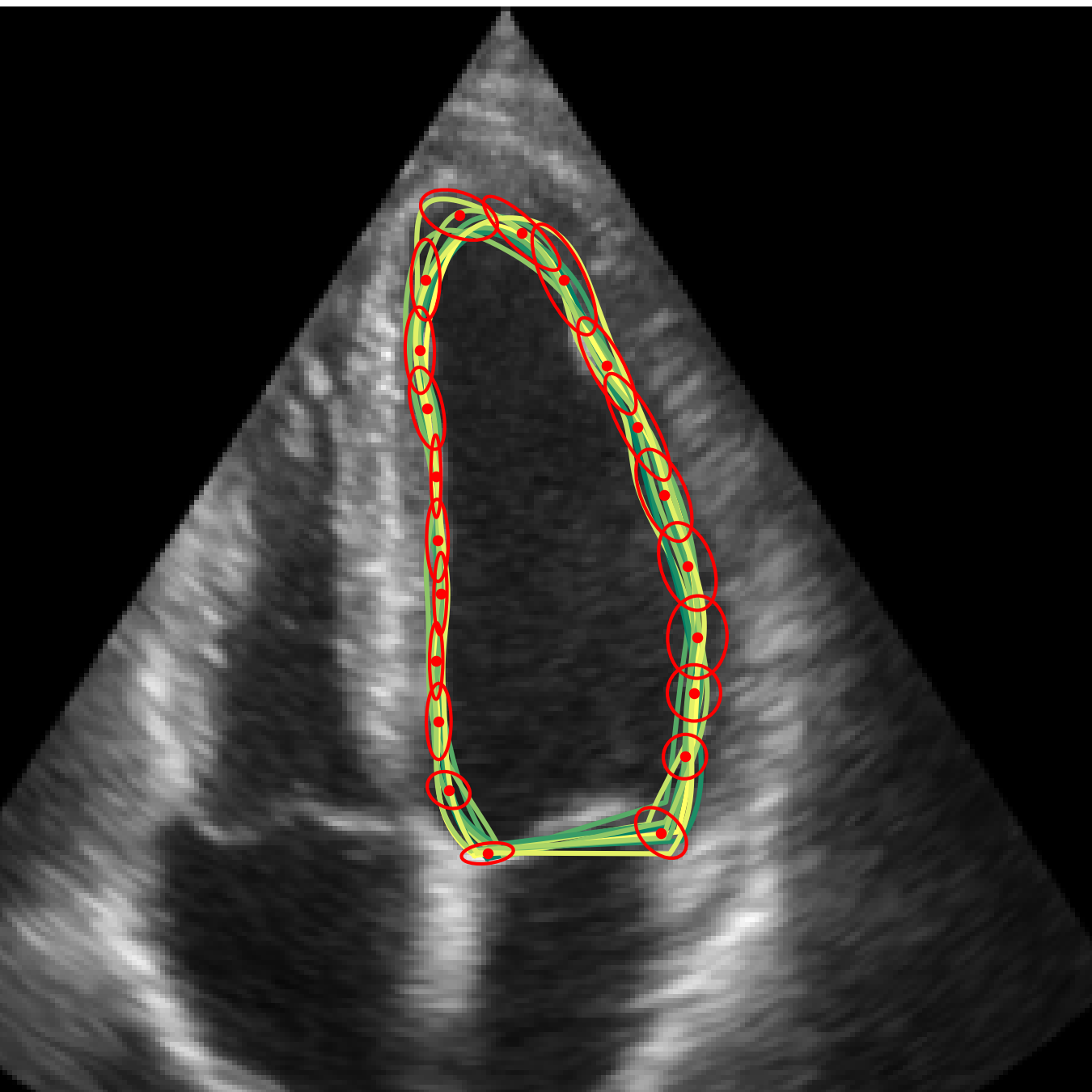}
  }\hfill
  \subfloat[\label{fig:orchid2}]{
    \includegraphics[width=0.23\textwidth]{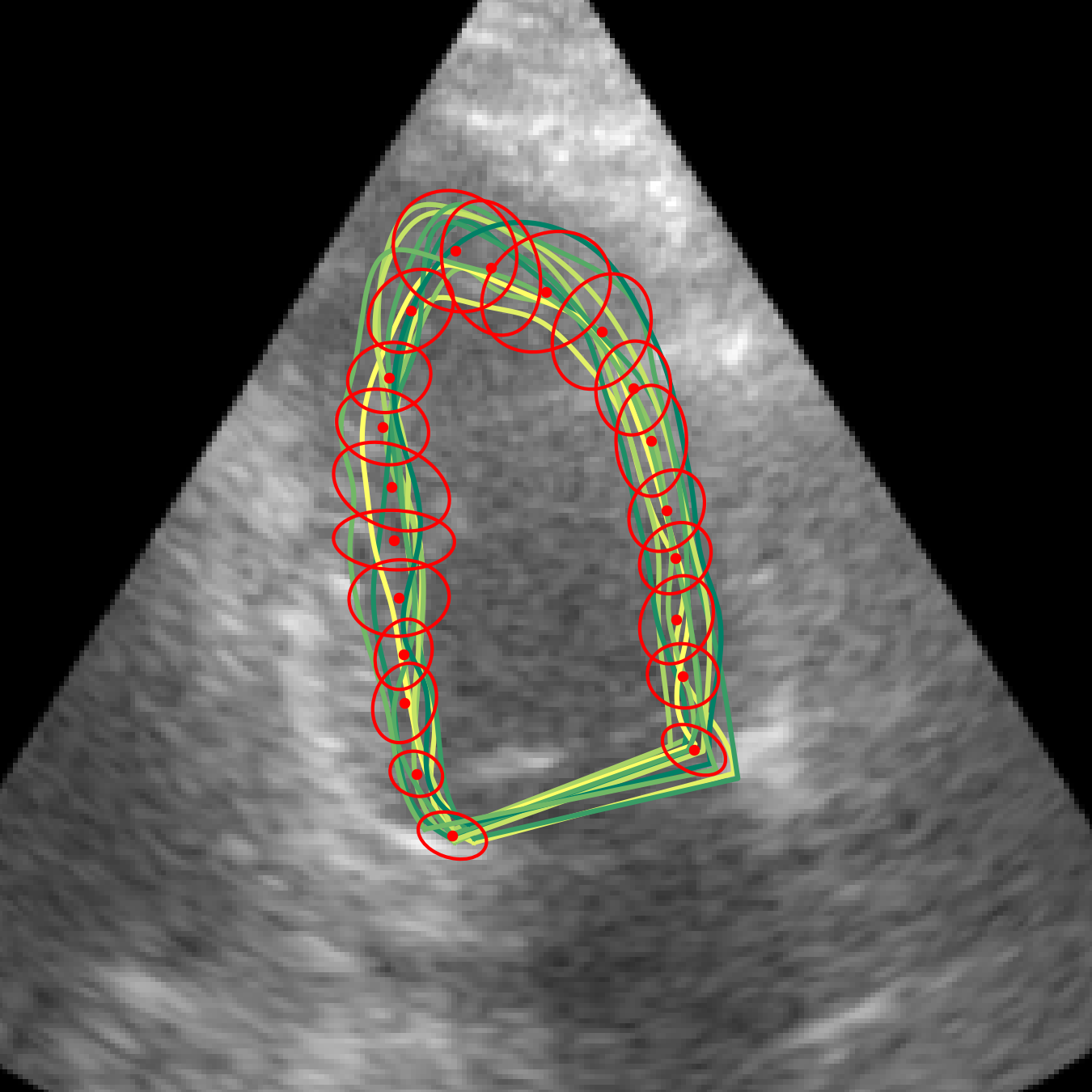}
  }\hfill
  \subfloat[\label{fig:orchid3}]{
    \includegraphics[width=0.23\textwidth]{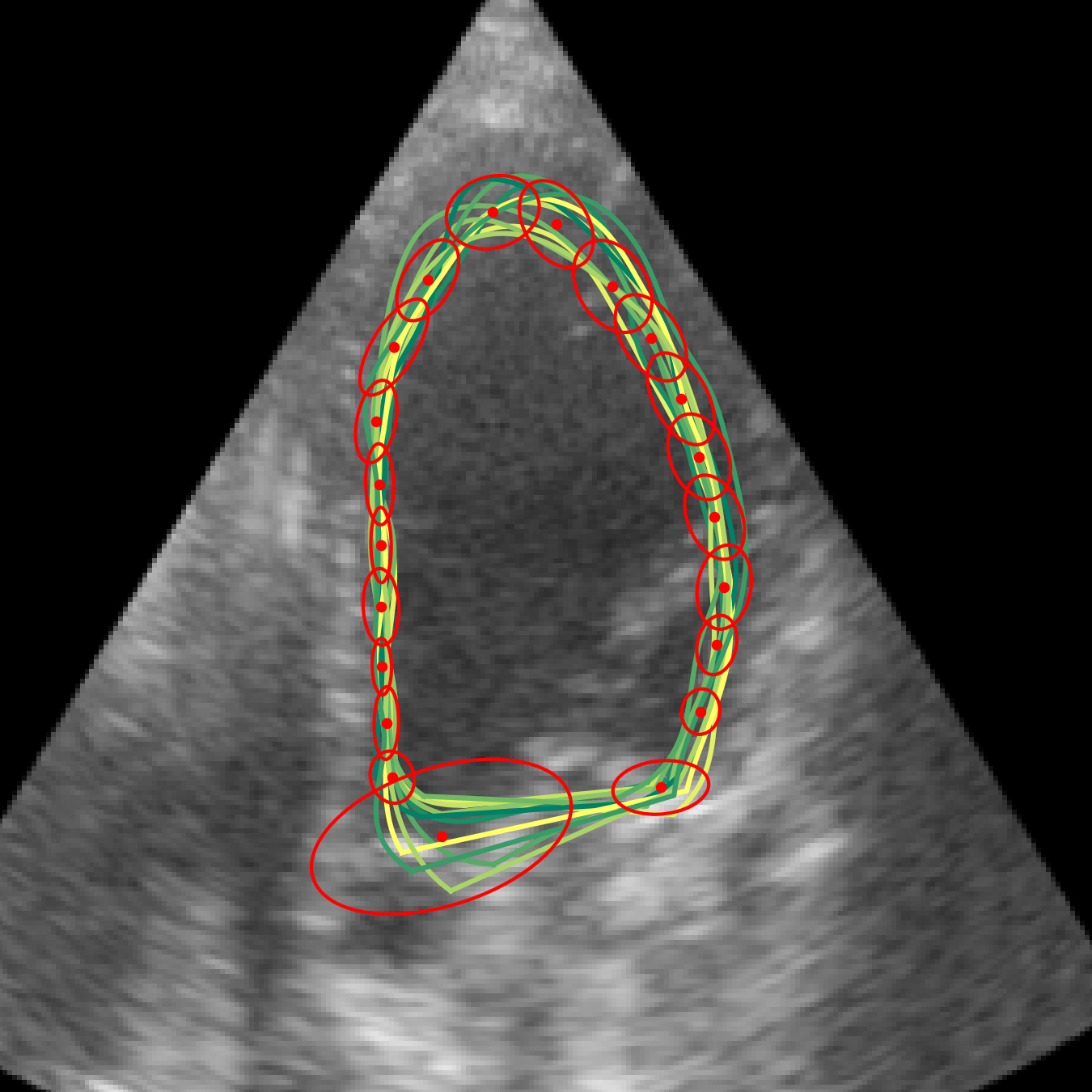}
  }
  \hfill
  \subfloat[\label{fig:orchid4}]{
    \includegraphics[width=0.23\textwidth]{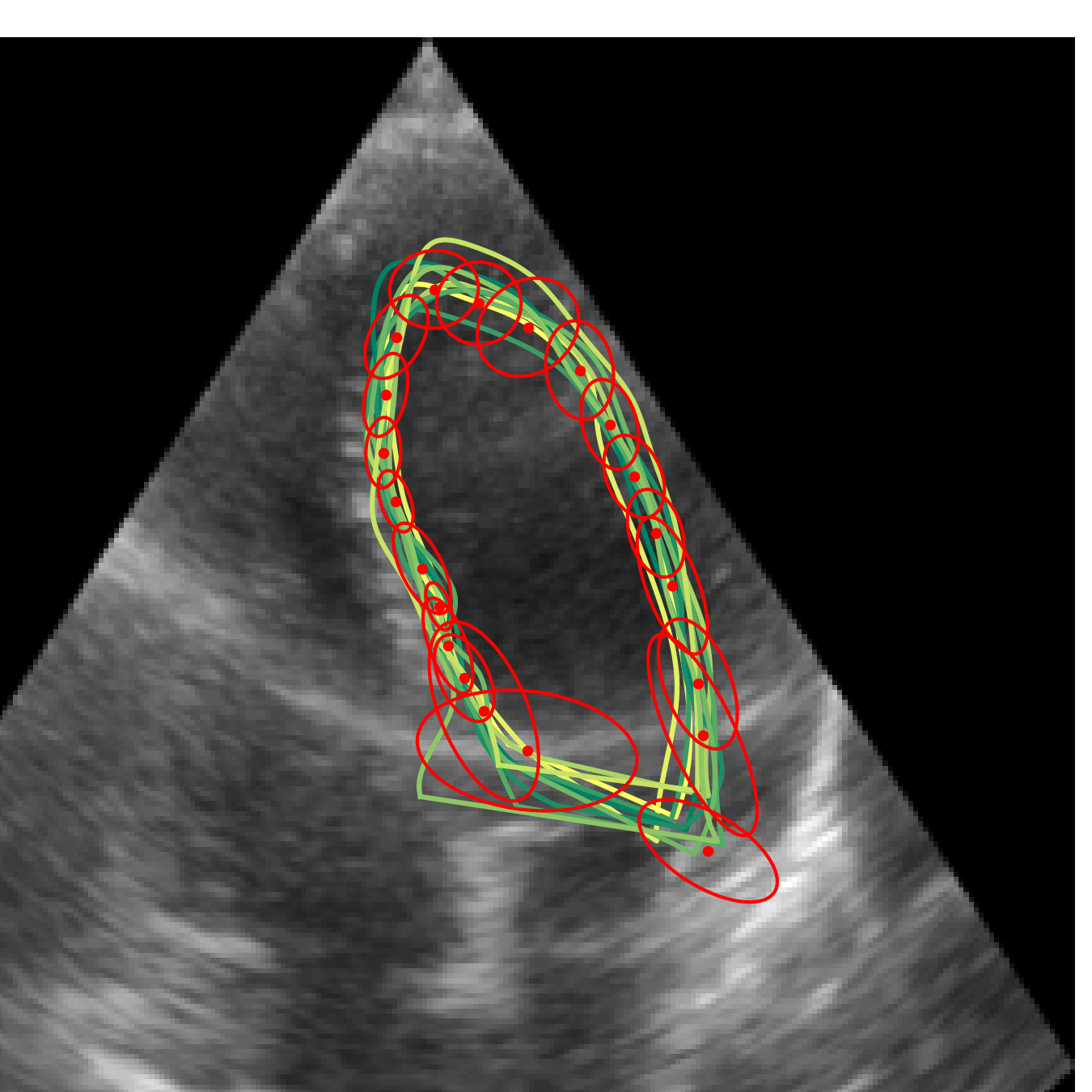}
  }
  \caption{\added{Samples from the private French test set. (a) Sample from class 2 with good quality and high contrast. The uncertainty is highest along the contour direction, providing little uncertainty during the sampling process. (b) Sample from class 3 with low contrast with high uncertainty perpendicularly to the contour. (c) Sample from class 4 with noisy acquisition as the aorta is visible, causing high uncertainty at the basal valve. (d) Sample from class 4, with reverberation present close to the valves making them less visible and increasing uncertainty in this area.}}
\label{fig:sampling_method}
\end{figure*}

\subsection{\added{Extension to other chambers}}
\label{sec:la}
\begin{figure}[tp]
\centering
\includegraphics[width=0.65\linewidth]{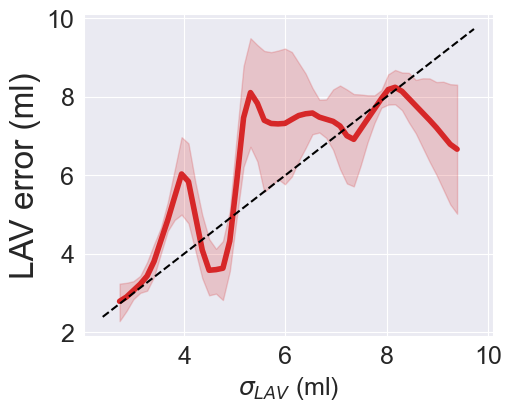}
\caption{\added{Calibration of CASUS-t MC-D for the left atrium volume on 54 volumes.}}
\label{fig:la_cal}
\end{figure}

\begin{figure}[tp]
  \centering
    \subfloat[\label{fig:atrium1}]{
        \includegraphics[width=0.47\columnwidth]{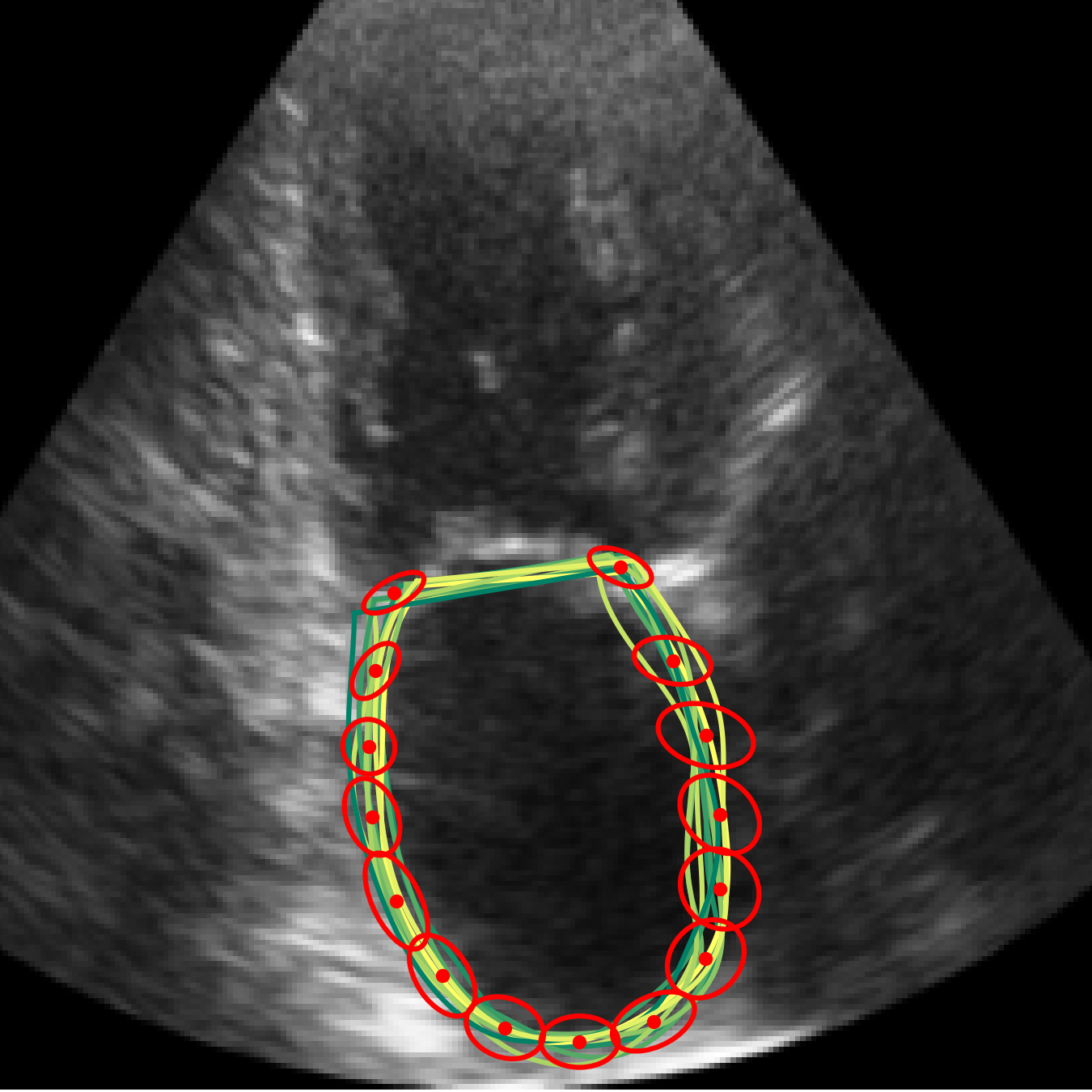}
    }
    \hfill
    \subfloat[\label{fig:atrium2}]{
        \includegraphics[width=0.47\columnwidth]{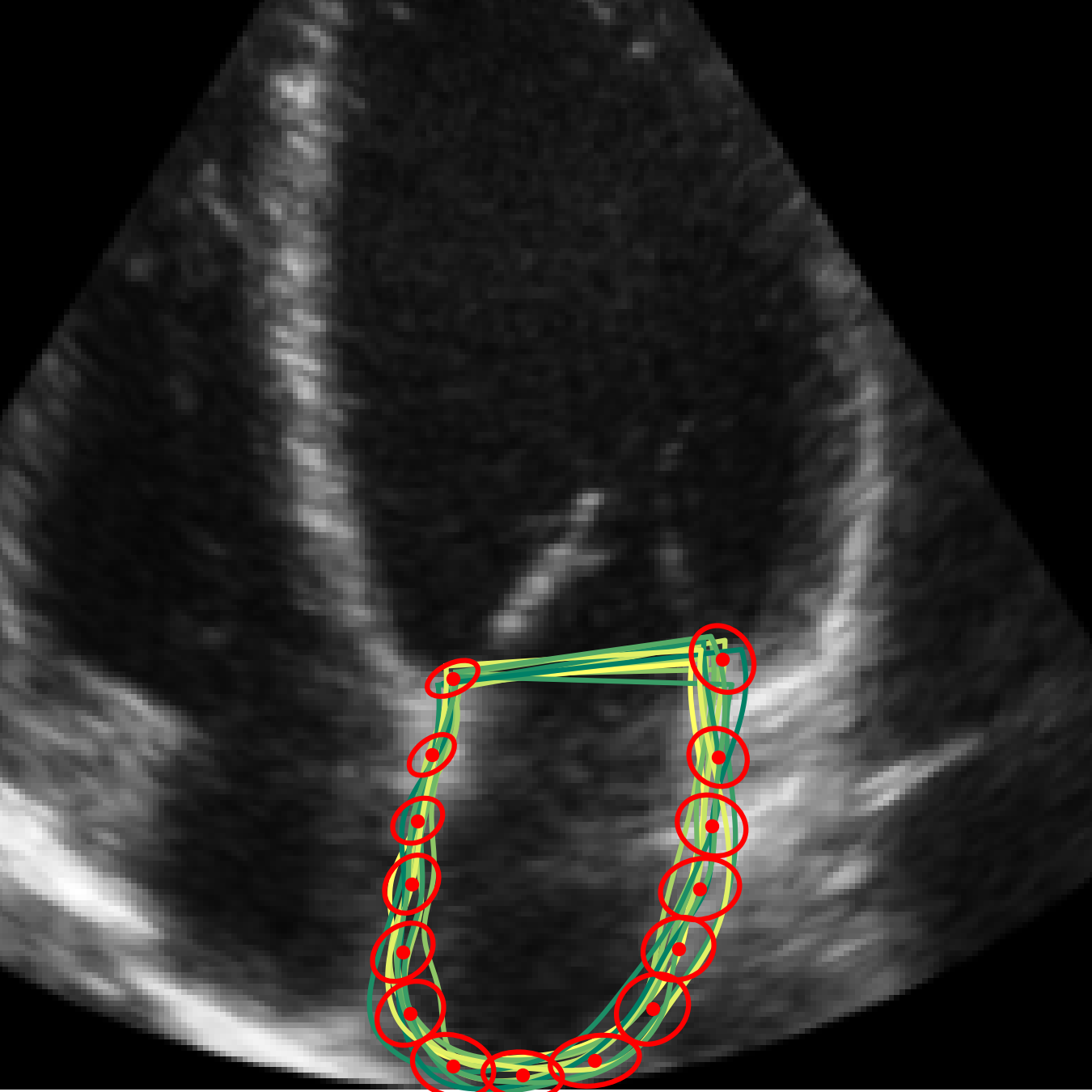}
    }
  \caption{\added{Samples of our method on the left atrium.}}
\label{fig:atrium}
\end{figure}

\added{
The proposed method is not limited to the left ventricle. Our method can be applied to any structure that a succession of points can define. The left atrium (LA) is a ideal structure for this.  As a proof of concept, we therefore explored the extension of our method to LA segmentation using the CAMUS dataset, which also provides LA annotations.
}
\added{
Since CAMUS acquisitions are optimized for LV visualization, some images do not fully capture the LA, preventing reliable contour definition. These samples were excluded from both the training and test sets, resulting in 1023 training images (64\% of the original dataset) and 27 testing patients where both views and both instances contain the full atrium. We trained five models on this subset. Our method achieved a Dice score of $.923\pm0.003$, consistent with state-of-the-art results~\cite{Leclerc19}.
}

\added{
We further computed LA volume using the biplane Simpson method \cite{KEBED2019191_LA} and evaluated the calibration of our uncertainty estimation, with the corresponding calibration curve presented in \Cref{fig:la_cal}. LA volume calibration (UCE = $8.87\pm1.71$ ml) was broadly comparable to that observed for LV volume in the main analysis. Qualitative examples of LA segmentation are shown in \Cref{fig:atrium}.
}

\added{
These preliminary results suggest that the proposed framework can be extended to LA segmentation, although further dedicated calibration and validation are required before clinical application.
}

\section{Conclusion}

In this work, we presented a novel framework for uncertainty propagation in the estimation of clinical metrics from segmentations derived from echocardiographic images. This framework provides a reliable uncertainty estimation that can combine both aleatoric and epistemic uncertainty. Although this adaptable framework accommodates any uncertainty method capable of generating samples, our findings highlight that a contour-based approach to aleatoric uncertainty, combined with shape uncertainty modeling, delivers optimal results in clinical metric calibration. This approach is further improved by integrating temporal consistency in the sampling process. 

By producing uncertainty estimates at every step of the clinical metric–estimation pipeline, this work supports automated quality control and improves the robustness of diagnostic and prognostic models. Presenting predictions as ranges, rather than fixed point estimates, offers clinicians more informative and reliable decision-support data.

Although we \added{mainly} applied our method to the left ventricle, our method could be applied to any shape defined by a succession of points, such as the other heart chambers \added{as shown in \Cref{sec:la}}. Future work could address the aspect of interpretability of this pipeline as well as the user interactions on contour points in clinical settings.

One limitation of our work is the incompatibility between the variability required to achieve clinical metric calibration and pixel-wise uncertainty calibration. The increased variability in sample generation needed for modeling clinical uncertainty leads to under-confident pixel-wise estimations. CASUS and SSN exhibit better calibration on clinical metrics but show under-confidence in pixel-wise calibration. Alternative methods for generating uncertainty maps from sample sets may further enhance calibration. \added{Moreover, the results highlight that generalization remains more challenging in specific subgroups, particularly for rare or atypical pathologies such as HCM-CA and for the smallest left ventricular volumes. Reliable uncertainty calibration in these settings remains an open challenge and warrants further investigation on larger, disease-annotated cohorts. Finally, it is worth noting that the datasets used in the current evaluation include a limited number of labeled pathologies, with some notably absent, such as valvular disease. This represents a remaining limitation of the generalizability analysis and should be addressed in future studies.}

\bibliographystyle{IEEEtran}
\bibliography{ref}
\clearpage
\listofchanges


 





\end{document}